\documentclass{gensi}

\usepackage{amsmath,amssymb}
\usepackage{array}
\usepackage{tabularx}
\usepackage{xurl}
\usepackage{algorithm}
\usepackage{algpseudocode}

\usepackage{amsmath,amsfonts,bm}

\def\eqref#1{equation~\ref{#1}}
\def\Eqref#1{Equation~\ref{#1}}
\def\1{\bm{1}}

\DeclareMathAlphabet{\mathsfit}{\encodingdefault}{\sfdefault}{m}{sl}
\SetMathAlphabet{\mathsfit}{bold}{\encodingdefault}{\sfdefault}{bx}{n}

\definecolor{RecipeLink}{RGB}{20,90,145}
\definecolor{RouteGray}{RGB}{110,110,110}
\newcommand{\hflink}[2]{\href{#1}{\textcolor{RecipeLink}{#2}}}
\newcommand{\routegrey}[1]{\textcolor{RouteGray}{#1}}

\title{Open-MOPD: Diagnosing and Fixing Capability Imbalance in\\
Multi-Teacher On-Policy Distillation}

\author[*,\ddagger,1,2,3]{Huan-ang Gao}
\author[*,1,2,3]{Haohan Chi}
\author[1,2,3]{Yong Yan}
\author[1,2]{Shiyuan Feng}
\author[1,2]{Hanlin Wu}
\author[3]{Zheng Jiang}
\author[3]{Bingxiang He}
\author[1,2]{Wei-Ying Ma}
\author[1,2]{Ya-Qin Zhang}
\author[1,2,\dagger]{Hao Zhou}

\affiliation[1]{SIA-Lab of Tsinghua AIR and ByteDance Seed}
\affiliation[2]{Institute for AI Industry Research (AIR), Tsinghua University}
\affiliation[3]{Department of Computer Science and Technology, Tsinghua University}

\contribution[*]{Equal contribution}
\contribution[\ddagger]{Project Lead}
\contribution[\dagger]{Corresponding author}

\abstract{Multi-teacher on-policy distillation (M-OPD) has emerged as a promising paradigm for consolidating domain-specialized reinforcement learning (RL) experts into a single generalist student via dense, token-level reward supervision. Despite its practical success, the optimization dynamics governing multi-teacher capability integration remain poorly understood, and open, rigorously reproducible recipes are conspicuously lacking. In this work, we establish a controlled M-OPD benchmark on SmolLM3-3B-Base with oracle routing, isolating capability integration from routing ambiguity. Our investigation reveals a pronounced \textit{capability integration gap}: standard M-OPD recovers only $35.6\%$ of the improvement from mixed-domain SFT to RouteRL. Concise tasks such as instruction following suffer severe degradation and premature stagnation. Under this metric, token-level teacher disagreement does not appear to be the dominant bottleneck; the main failure is a severe misallocation of the token-level optimization budget. We identify three separable contributors to this imbalance: structural sequence-length disparities across domains, different convergence rates across domains, and reward staleness from stale student updates across repeated minibatches of a shared rollout. To resolve these imbalances, we introduce \textbf{Open-MOPD}, a principled framework incorporating token-share balancing, gap-aware dynamic budget allocation, and student reward refresh. Together, these mechanisms systematically restore cross-domain balance, elevating recovery from $35.6\%$ to $83.4\%$ of the improvement from mixed-domain SFT to RouteRL in one student model. We fully open-source our end-to-end post-training recipe, training trajectories, and evaluation suites on an 8$\times$A100-80GB academic setup.
\par\noindent
Project page: \url{https://bytedtsinghua-sia.github.io/Open-MOPD/}}

\begin{document}

\maketitle

\section{Introduction}
\label{sec:intro}

Reinforcement learning (RL) has established itself as a foundational pillar in large language model (LLM) post-training, excelling at cultivating specialized capabilities in domain-specific models~\citep{shao2024deepseekmath, yu2025dapo}. However, serving multiple specialized models in real-world applications is often computationally prohibitive, creating a strong demand for a single unified model that preserves these disparate strengths. To consolidate multiple specialists into a single student, multi-teacher on-policy distillation (M-OPD) has emerged as an appealing paradigm; each input prompt is routed to a corresponding domain expert, which provides dense per-token reward supervision over the trajectories generated by the student policy~\citep{li2026rethinking, ma2026mopd, bercovich2025llama}. Despite the increasing adoption of multi-teacher distillation in industrial pipelines, the fundamental mechanisms governing multi-domain capability integration remain poorly understood, and the research community still lacks an open and rigorously reproducible recipe.

To isolate and examine the core dynamics of multi-teacher consolidation, we construct a fully controlled M-OPD experimental framework initialized from SmolLM3-3B-Base. The pipeline incorporates a three-domain mixed-SFT initialization, three domain-expert RL teachers (spanning mathematics, coding, and general instruction following), and oracle routing based on ground-truth domain labels. Using ground-truth domain labels, we isolate capability integration from routing errors. This design operates at a parameter scale that balances two criteria (Section~\ref{sec:recipe:model}), and the entire pipeline and ablation ladder are reproducible on an 8$\times$A100-80GB academic setup, while the model remains capable of long-horizon reasoning.

\begin{figure}[htbp]
\centering
\includegraphics[width=\linewidth]{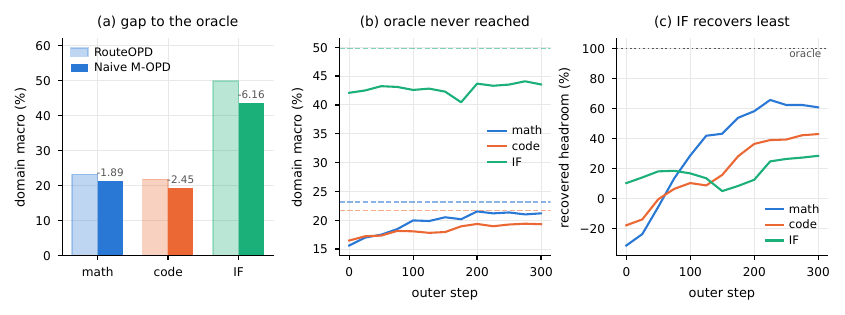}
\vspace{-4mm}
\caption{The integration gap is real and strongly asymmetric.
(a) Per-domain macro-averages of Naive M-OPD and RouteOPD, where the number above each bar indicates the gap in that domain;
(b) per-domain validation trajectories of Naive M-OPD, with dashed lines denoting the RouteOPD reference for the corresponding domain;
(c) recovered theoretical headroom relative to RouteRL,
$(\text{current}-\pi_{\mathrm{mixsft}})/(\text{RouteRL}-\pi_{\mathrm{mixsft}})$.
Instruction following (IF) exhibits the largest absolute gap and the lowest recovered headroom, and is the earliest domain to stop improving.}
\label{fig:gap_zh}
\end{figure}

Our empirical investigation reveals that perfect routing is far from sufficient for successful capability integration. Under a standardized six-benchmark evaluation protocol, naive M-OPD reaches an average score of $28.05$, while distilling each domain on its own reaches $31.55$. We use RouteOPD to measure the deployment-time integration gap. In particular, concise instruction-following (IF) tasks fall $6.16$ points below their RouteOPD reference ($3.3\times$ the degradation observed in mathematics) and plateau earliest during training.

Through systematic per-domain measurement, we locate this integration gap in the unmanaged allocation of the token-level optimization budget rather than catastrophic gradient conflict (Section~\ref{sec:diagnosis}). Because on-policy distillation objectives aggregate loss values across tokens, the actual optimization share received by each domain is governed by its gradient token volume rather than its prompt frequency. Consequently, concise IF responses contribute a negligible fraction of gradient tokens despite receiving a balanced prompt allocation, accounting for $20.3\%$ of the input prompts but only $0.99\%$ of the gradient tokens. Furthermore, even if the raw token budget is equalized initially, it drifts apart within tens of steps because the supervisory reward signals diminish at uneven rates as the student converges toward different teachers at varying speeds. This imbalance is compounded by standard multi-step rollout reuse, which leaves student-dependent reward components stale across successive gradient updates.

To address these orthogonal failure modes, we introduce Open-MOPD, a principled framework that targets each distortion with a dedicated mechanism (Section~\ref{sec:method_zh}). First, \textit{token-share balancing} decouples gradient budget allocation from sequence length, establishing a weighted token-share target rather than a fixed per-batch quota. Second, \textit{gap-aware allocation} dynamically steers the optimization budget toward domains with the largest remaining student--teacher gap. This prevents the training collapse observed with naive reward normalization at step 74, where budget is wastefully funneled into already-converged domains. Third, \textit{student reward refresh} recomputes student log-probabilities before each gradient step while caching teacher states, eliminating sample staleness with negligible overhead~\citep{kang2026asyncopd}. In cumulative ablation experiments, Open-MOPD raises recovery from $35.6\%$ to $83.4\%$ in one student model.

Our principal contributions are summarized below.
\begin{itemize}
\item \textbf{Empirical Diagnosis.} Using an oracle-routed testbed that isolates integration dynamics from routing errors, we identify and characterize the \textit{capability integration gap} in multi-teacher on-policy distillation, revealing pronounced cross-domain performance asymmetry and premature stagnation in concise tasks (Section~\ref{sec:diagnosis}).

\item \textbf{Mechanistic Decomposition.} Token-level teacher disagreement under our metric is unlikely to be the dominant bottleneck. We instead identify three separable contributors to token-level optimization budget distortion: sequence-length disparities, different convergence rates, and reward staleness from stale student updates across repeated minibatches of a shared rollout (Section~\ref{sec:method_zh}).

\item \textbf{Principled Methodology.} We propose Open-MOPD, an optimization framework combining token-share balancing, gap-aware dynamic budget allocation, and student reward refresh, which resolves domain imbalances and recovers most of the available headroom in one student model (Section~\ref{sec:method_zh}, Section~\ref{sec:ablation}).

\item \textbf{Open-Source Recipe.} We fully open-source our end-to-end post-training pipeline, model checkpoints, and evaluation suites on a reproducible academic compute budget (Appendix~\ref{app:recipe}).
\end{itemize}

\section{Open-MOPD: An Open Recipe for Multi-Teacher Capability Integration}
\label{sec:recipe}

Our goal is to integrate the capabilities of several domain experts into one unified
student model. Let the set of domains be $\mathcal{D}$. On our base model,
$\mathcal{D}=\{\mathrm{math},\mathrm{code},\mathrm{instruction\_following}\,(\mathrm{IF})\}$.
Each domain has an expert teacher $\pi_{\phi_d}$ trained with reinforcement learning, and the student
model is $\pi_\theta$. Each training sample carries a domain label d(x), allowing a single student model to receive supervision signals from the corresponding domain expert across three distinct tasks: mathematics, code generation, and instruction following. With the training framework fixed, Open-MOPD aims to identify a recipe that stably transfers the capabilities of all three experts into a single student model.

\subsection{The Multi-Teacher On-Policy Distillation Objective}
\label{sec:recipe:objective}

For a prompt $x$, the student first samples a response
$y\sim\pi_\theta(\cdot\mid x)$ from the current policy. The teacher evaluates student rollouts via a single prefill pass to generate the teacher distribution. Write the routed teacher
as $\pi_{\phi_{d(x)}}$. At position $t$, we take the top-$k$ token set of the student distribution
\begin{equation}
\label{eq:topk_set_zh}
\mathcal{S}_t
=
\mathrm{TopK}_k\!\big(\pi_\theta(\cdot\mid x,y_{<t})\big),
\end{equation}
with $k=16$ by default. For each $v\in\mathcal{S}_t$, we first define the teacher--student
log-probability difference
\begin{equation}
\label{eq:opd_gap_zh}
\delta_t(v)
=
\mathrm{sg}\!\Big[
\log\pi_{\phi_{d(x)}}(v\mid x,y_{<t})
-
\log\pi_\theta(v\mid x,y_{<t})
\Big],
\end{equation}
where $\mathrm{sg}[\cdot]$ denotes stop-gradient; this is the token-level advantage signal of dense
distillation. It is then aggregated with softmax weights over $\mathcal{S}_t$ into a dense reward
\begin{equation}
\label{eq:opd_reward_zh}
r_t(v)
=
\delta_t(v)\cdot\tilde\pi_\theta(v\mid x,y_{<t}),
\qquad
\tilde\pi_\theta(v\mid x,y_{<t})
=
\mathrm{softmax}_{u\in\mathcal{S}_t}\!\big[
\log\pi_\theta(u\mid x,y_{<t})
\big](v),
\end{equation}
and tokens outside $\mathcal{S}_t$ contribute zero. The position-level reward is
\begin{equation}
\label{eq:opd_pos_zh}
r_t
=
\sum_{v\in\mathcal{S}_t}r_t(v).
\end{equation}
With no critic, $r_t$ is placed directly in the advantage slot of PPO. Intuitively, $\delta_t(v)>0$
means that the teacher assigns a higher probability to token $v$ than the student does; $\tilde\pi_\theta$ only
determines the share each token inside the top-$k$ contributes to the position-level reward $r_t$.
Because the student generates every response, the supervision distribution follows the student's
current policy. This is the key difference between on-policy distillation and supervised fine-tuning
on offline teacher trajectories, whose fixed data distribution can create a distribution shift. For
this reason, Section~\ref{sec:diagnosis} starts from this minimal version, which adds no further
training mechanisms and keeps the effects of later changes easier to identify.

\subsection{Base Model Selection}
\label{sec:recipe:model}

The open recipe uses SmolLM3-3B-Base as its base model. It is a fully open 3B decoder-only model
whose pre-training covers web, math and code data and which is trained at a 64K
context~\citep{smollm3}. This choice balances compute cost with the feasibility of running the full
recipe and repeating its ablations.

\paragraph{Experimental feasibility.}
Multi-teacher OPD needs student generation, one or more teacher forwards, long-response training and
per-domain validation at once; a full reproduction also includes mixed-domain SFT, three domain RL
teachers and the final multi-teacher distillation. If the base model started at 7B or larger, that
chain would be hard to close within a budget on which an ordinary academic team can ablate
repeatedly: the GPU hours taken by a single experiment rise, and the repeated runs needed for
mechanism comparisons become unaffordable. Open-MOPD therefore takes ``runnable on a single
8$\times$A100-80GB node'' as a recipe constraint, so that the end-to-end pipeline and the mechanism
ablations can be executed repeatedly with limited computational resources.

\paragraph{Model capacity.}
The base model must also have enough capacity. A model that is too small lets the response
length limit truncate long-chain supervision before it enters a usable trajectory; for a failed
experiment, it is difficult to determine whether the failure is caused by the algorithm or training
configuration, or by the model's limited ability to discover and learn solutions within the response
length limit. We observed this failure mode earlier with the smaller Qwen3-1.7B-Base as base model:
across five epochs of SFT on the same OpenR1-Math-93k, its best AIME24 result was only
7.08\%, and even with a 31K generation budget the truncation rate stayed at 69.17--80.42\%.
Qwen2.5-7B-Base on the same data reached 31.25\%, with the truncation rate down to 12.92\%
(Appendix~\ref{app:recipe:base_model}). When most responses are truncated, a failed run reveals little
about whether the training method itself is effective.
SmolLM3-3B-Base is large enough to learn from the long responses used in our math and code SFT,
which has a 32,768-token limit. This lets us measure the integration gap and run the ablations
without response truncation dominating the results.

\subsection{The End-to-End Training Recipe}
\label{sec:recipe:pipeline}

The Open-MOPD recipe has three stages. The teachers in Stage II and the student in Stage III are
initialized from the same mixed-domain SFT model.
Table~\ref{tab:recipe_pipeline_zh} summarizes the base model, the data, and the evaluation settings of the recipe.

\begin{table}[ht]
\caption{\textbf{The Open-MOPD pipeline.} Starting from a mixed-domain SFT checkpoint ($\pi_{\mathrm{mixsft}}$), domain-specific RL produces three teacher models $\pi_{\phi_d}$. Multi-teacher OPD then trains the student model $\pi_\theta$ from the initial checkpoint. All datasets are publicly released for end-to-end reproducibility.}
\label{tab:recipe_pipeline_zh}
\centering
\small
\renewcommand{\arraystretch}{1.25}
\setlength{\tabcolsep}{0pt}
% Use ordinary p-columns here.  They are more robust across template versions
% than X columns, while the final column still absorbs the remaining width.
\begin{tabular}{@{}>{\raggedright\arraybackslash}p{1.45cm}
                @{\hspace{0.75em}}
                >{\raggedright\arraybackslash}p{5.7cm}
                @{\hspace{0.75em}}
                >{\raggedright\arraybackslash}p{\dimexpr\linewidth-1.45cm-5.7cm-1.5em\relax}@{}}
\toprule
Stage & Data & Key settings and products \\
\midrule
Base model &
\hflink{https://huggingface.co/HuggingFaceTB/SmolLM3-3B-Base}{SmolLM3-3B-Base}
&
3B; 1.7B fails to close long trajectories, while 7B is too costly for repeated ablations
(Appendix~\ref{app:recipe:base_model})
\\
\midrule
I\newline Mixed-domain SFT &
Math:
\hflink{https://huggingface.co/datasets/open-r1/OpenR1-Math-220k}{OpenR1-Math-93k}
\newline
Code:
\hflink{https://huggingface.co/datasets/nvidia/OpenCodeReasoning}{OCR-50k}
(sampled from the full set)
\newline
IF:
\hflink{https://huggingface.co/datasets/nvidia/Llama-Nemotron-Post-Training-Dataset}{Instruction-Nemotron}
&
Four epochs; sequence limit 32,768; approximately balanced by response tokens (37/28/35)
\newline
product $\pi_{\mathrm{mixsft}}$
\\
\midrule
II\newline Domain RL teachers &
Math:
\hflink{https://huggingface.co/datasets/BytedTsinghua-SIA/DAPO-Math-17k}{DAPO-Math-17k}
\newline
Code:
\hflink{https://huggingface.co/datasets/agentica-org/DeepCoder-Preview-Dataset}{DeepScaler-24k}
(LCB-decontaminated)
\newline
IF:
\hflink{https://huggingface.co/datasets/nvidia/Nemotron-Cascade-2-RL-data}{Nemotron-IF-RL-46k}
&
Independent RL per domain, no domain mixing; all forked from $\pi_{\mathrm{mixsft}}$
\newline
products $\pi_{\phi_{\mathrm{math}}},\pi_{\phi_{\mathrm{code}}},\pi_{\phi_{\mathrm{IF}}}$
\\
\midrule
III\newline Multi-teacher OPD &
The union of the three domains' RL prompts above; each prompt is hard-routed to its teacher by domain
label
&
Response limit 16K for math/code and 2K for IF; top-$k{=}16$; the student is initialized from
$\pi_{\mathrm{mixsft}}$
\newline
product $\pi_\theta$
\\
\midrule
Evaluation &
Math:
\hflink{https://huggingface.co/datasets/HuggingFaceH4/aime_2024}{AIME24}/
\hflink{https://huggingface.co/datasets/math-ai/aime25}{AIME25}
\newline
Code:
\hflink{https://huggingface.co/datasets/livecodebench/code_generation_lite}{LiveCodeBench}
v5/v6
\newline
IF:
\hflink{https://huggingface.co/datasets/google/IFEval}{IFEval}+
\hflink{https://huggingface.co/datasets/allenai/IFBench_test}{IFBench$_{\mathrm{test}}$}
&
mean@64 / mean@10 / mean@1; averaged evenly across the datasets within a domain first, then
macro-averaged over the three domains
\\
\bottomrule
\end{tabular}
\end{table}

\paragraph{Stage I: mixed-domain SFT.}
Starting from SmolLM3-3B-Base, we balance the sample count of each domain by response token count and
then run four epochs of supervised fine-tuning on the mixed data, giving the shared student checkpoint
$\pi_{\mathrm{mixsft}}$.
The data mixture is: math OpenR1-Math-93k (93,733 examples), code OCR-50k sampled from the full
OpenCodeReasoning set (50,000 examples), and instruction-following Instruction-Nemotron aligned
(820,039 examples); estimated by response tokens, the shares are approximately 37\% math,
28\% code and 35\% IF. The sequence length limit is 32,768.

\paragraph{Stage II: per-domain RL teachers.}
Three teachers are initialized separately from $\pi_{\mathrm{mixsft}}$, each running RL only on the
verifiable reward of its own domain.
The math teacher is trained on DAPO-Math-17k; the code teacher is trained on DeepScaler-24k (the
LiveCodeBench-decontaminated version); the IF teacher is trained on Nemotron-IF-RL-46k. This stage gives
$\pi_{\phi_{\mathrm{math}}},\pi_{\phi_{\mathrm{code}}},\pi_{\phi_{\mathrm{IF}}}$.

\paragraph{Stage III: multi-teacher on-policy distillation.}
The student is also initialized from $\pi_{\mathrm{mixsft}}$. Training prompts come from the union of
the three domains and are sampled as a domain mixture. The student $\pi_\theta$ generates a response
for each sampled prompt. The domain label then selects the corresponding teacher for OPD using the
objective in Section~\ref{sec:recipe:objective}.
The response length limit is 16K for math and code and 2K for IF; by default the dense reward is
computed on the student top-$k$ ($k=16$).

\subsection{Evaluation Setup}
\label{sec:recipe:protocol}

For each domain, we use the best performance obtained by training a student with its domain
teacher alone as a reference. These three single-teacher students together form RouteOPD, our
deployment-time integration-gap reference. At evaluation time, the domain label selects the
corresponding student to generate the response. The performance gap between the unified
multi-teacher OPD (M-OPD) student and RouteOPD defines the \emph{integration gap}.

All main results use the same online verifier and the same six benchmarks.
Scoring is done per dataset first, then averaged simply over the datasets within a domain, and the
final total score is the simple average of the three domain averages.
The math domain uses AIME24 and AIME25 with $n=64$ and temperature $=0.6$, reporting accuracy
mean@64;
the instruction-following domain uses IFEval and IFBench$_{\mathrm{test}}$ (abbreviated IFB$_{\mathrm{test}}$ in tables) with $n=1$, reporting
accuracy mean@1;
the code domain uses LiveCodeBench v5 and v6 with $n=10$ and temperature $=1.0$, reporting accuracy
mean@10.

Table~\ref{tab:baseline_gap_zh} provides the reference results for SmolLM3-3B under this
evaluation protocol. It includes the base model, the four mixed-domain SFT checkpoints, the
single-domain RL and OPD results, and their domain-routed combinations RouteRL and RouteOPD.
It also includes the RFT, $\pi_{\mathrm{mixrl}}$, Parameter Merging, Naive M-OPD, and Open-MOPD
comparisons. The analysis and ablations below use this table as their common reference.

\begin{table}[t]
\caption{\textbf{Integration gap on SmolLM3-3B.}
Each domain reports its sub-dataset scores and macro-average. We define the total score as the
average of the three domain averages. The four SFT rows are checkpoints after successive epochs.
Single-domain RL and OPD report results only for their respective domains. Gray-shaded
\textit{RouteRL} and \textit{RouteOPD} use a separate model for each domain and select that model
using the domain label at evaluation time. They therefore cannot be deployed as one model. The
recovery rate is $\frac{\text{Row Total} - \text{SFT}}{\text{RouteRL} - \text{SFT}}$, using
$\pi_{\mathrm{mixsft}}$ (epoch 4) as the SFT baseline. In the \textit{Baselines} and
\textit{M-OPD} sections, the highest score in each column is boldfaced and the second highest is
underlined.}
\label{tab:baseline_gap_zh}
\centering
\scriptsize
\setlength{\tabcolsep}{2.6pt}
\resizebox{\linewidth}{!}{%
\begin{tabular}{@{}l ccc c ccc c cc c c c@{}}
\toprule
& \multicolumn{3}{c}{Math (avg@64)} && \multicolumn{3}{c}{Code (avg@10)} && \multicolumn{3}{c}{IF} & & \\
\cmidrule(lr){2-4}\cmidrule(lr){6-8}\cmidrule(lr){10-12}
Method & AIME24 & AIME25 & avg && LCBv5 & LCBv6 & avg && IFEval & IFB$_{\mathrm{test}}$ & avg & Total & Recovery rate \\
\midrule
SmolLM3-3B-Base
  & 2.08 & 1.72 & 1.90 && 3.41 & 6.17 & 4.79 && 16.08 & 13.00 & 14.54 & 7.08 & --- \\
\midrule
\multicolumn{14}{@{}l}{\textit{SFT}} \\
$\pi_{\mathrm{mixsft}}$ (e1)
  & 12.24 & 16.61 & 14.43 && 13.29 & 17.20 & 15.25 && 64.70 & 16.00 & 40.35 & 23.34 & --- \\
$\pi_{\mathrm{mixsft}}$ (e2)
  & 12.55 & 17.97 & 15.26 && 13.17 & 16.63 & 14.90 && 65.62 & 16.00 & 40.81 & 23.66 & --- \\
$\pi_{\mathrm{mixsft}}$ (e3)
  & 15.10 & 19.53 & 17.32 && 14.97 & 18.57 & 16.77 && 66.36 & 18.00 & 42.18 & 25.42 & --- \\
$\pi_{\mathrm{mixsft}}$ (e4)
  & 15.63 & 20.26 & 17.95 && 15.99 & 19.20 & 17.60 && 66.91 & 16.00 & 41.46 & 25.67 & --- \\
\midrule
\multicolumn{14}{@{}l}{\textit{RL}} \\
$\pi_{\phi_{\mathrm{math}}}$
  & 23.65 & 24.84 & 24.24 && --- & --- & --- && --- & --- & --- & --- & --- \\
$\pi_{\phi_{\mathrm{code}}}$
  & --- & --- & --- && 22.16 & 21.31 & 21.73 && --- & --- & --- & --- & --- \\
$\pi_{\phi_{\mathrm{IF}}}$
  & --- & --- & --- && --- & --- & --- && 74.49 & 27.67 & 51.08 & --- & --- \\
\routegrey{RouteRL}
  & \routegrey{23.65} & \routegrey{24.84} & \routegrey{24.24}
  && \routegrey{22.16} & \routegrey{21.31} & \routegrey{21.73}
  && \routegrey{74.49} & \routegrey{27.67} & \routegrey{51.08}
  & \routegrey{32.35} & \routegrey{100\%} \\
\midrule
\multicolumn{14}{@{}l}{\textit{OPD}} \\
Math-OPD
  & 22.34 & 23.96 & 23.15 && --- & --- & --- && --- & --- & --- & --- & --- \\
Code-OPD
  & --- & --- & --- && 22.28 & 21.14 & 21.71 && --- & --- & --- & --- & --- \\
IF-OPD
  & --- & --- & --- && --- & --- & --- && 75.60 & 24.00 & 49.80 & --- & --- \\
\routegrey{RouteOPD}
  & \routegrey{22.34} & \routegrey{23.96} & \routegrey{23.15}
  && \routegrey{22.28} & \routegrey{21.14} & \routegrey{21.71}
  && \routegrey{75.60} & \routegrey{24.00} & \routegrey{49.80}
  & \routegrey{31.55} & \routegrey{88.0\%} \\
\midrule
\multicolumn{14}{@{}l}{\textit{Baselines}} \\
RFT
  & \textbf{22.97} & \textbf{23.91} & \textbf{23.44}
  && 18.98 & 19.43 & 19.21
  && 55.08 & 18.67 & 36.87
  & 26.51 & 12.6\% \\
$\pi_{\mathrm{mixrl}}$
  & 21.15 & 22.14 & 21.64
  && 16.59 & 20.97 & 18.78
  && 70.24 & \underline{22.67} & 46.45
  & 28.96 & 49.3\% \\
ParamMerge-Avg
  & 18.91 & 20.99 & 19.95
  && 18.38 & 21.20 & 19.79
  && 70.06 & 17.67 & 43.86
  & 27.87 & 32.9\% \\
ParamMerge-TA
  & 21.93 & 22.76 & 22.34
  && \textbf{21.74} & \textbf{23.14} & \textbf{22.44}
  && \underline{71.53} & 21.67 & \underline{46.60}
  & \underline{30.46} & \underline{71.7\%} \\
\midrule
\multicolumn{14}{@{}l}{\textit{M-OPD}} \\
Naive M-OPD
  & 20.92 & 21.60 & 21.26
  && 17.53 & 20.99 & 19.26
  && 68.61 & 18.67 & 43.64
  & 28.05 & 35.6\% \\
Open-MOPD (Ours)
  & \underline{21.98} & \underline{22.86} & \underline{22.42}
  && \underline{20.84} & \underline{22.63} & \underline{21.73}
  && \textbf{74.49} & \textbf{24.67} & \textbf{49.58}
  & \textbf{31.24} & \textbf{83.4\%} \\
\bottomrule
\end{tabular}%
}
\end{table}
\FloatBarrier

\section{Diagnosing the Multi-Teacher Integration Gap}
\label{sec:diagnosis}

The recipe in Section~\ref{sec:recipe} provides an open baseline for multi-teacher OPD. It
combines mixed-domain SFT, one teacher for each domain, hard routing by domain label, and
on-policy distillation on the student's own responses. This setup gives all domains the correct
teacher signal while training one shared student. However, the shared student still falls short of
the domain-specific teachers. This section studies why the shared student does not retain all of
the capabilities learned by the domain teachers. We first establish the integration gap with a
simple baseline, then test teacher conflict as a possible cause, and finally measure how much
training signal each domain receives and how this signal changes during training.

\subsection{A baseline reveals the integration gap}
\label{sec:diagnosis:gap}

In Table~\ref{tab:baseline_gap_zh}, Naive M-OPD lifts $\pi_{\mathrm{mixsft}}$ from 25.67 to 28.05,
which shows that the multi-teacher signal is useful. Its score is 0.91 points below
$\pi_{\mathrm{mixrl}}$ and 3.50 points below RouteOPD.
RouteOPD reaches 31.55, showing that the teachers, the student initialization, and the single-teacher
distillation objective are sufficient to learn these capabilities. The score drops when the three
teacher signals are naively combined in one student.

Figure~\ref{fig:gap_zh} shows that the gap is uneven across domains. IF falls 6.16 points
below its RouteOPD reference, which is 3.3 times the math gap of 1.89 points. It is also the
main source of the total integration gap (Figure~\ref{fig:gap_zh}a).
Along the training trajectory (Figure~\ref{fig:gap_zh}b), none of the three domains reaches its corresponding RouteOPD reference within 300 steps. Figure~\ref{fig:gap_zh}c reports the fraction of each domain's gap to that reference that has been closed. Over the $[200,300]$-step window, math closes 64\% of its gap, code 40\%, and IF only 26\%. IF is also the only domain whose score decreases during the middle of training, falling by 11\% over the $[100,200]$ window.

\subsection{Testing teacher conflict}
\label{sec:diagnosis:conflict}

A natural hypothesis is that teachers of different domains give mutually contradictory token
preferences on the same student trajectory: although the math, code and IF teachers are each
selected on their own domain only, they share a large number of formatting words, connectives and
reasoning templates. If the teachers disagree on these common tokens, multi-task OPD can push the
gradients in different directions and interfere with the shared parameters. To test this hypothesis,
multi-teacher on-policy distillation lets all three teachers score the same context for every token
sampled by the student. We define
\begin{equation}
\label{eq:teacher_spread_zh}
c_t
=\max_{d\in\mathcal{D}}\log\pi_{\phi_d}(y_t\mid x,y_{<t})
-\min_{d\in\mathcal{D}}\log\pi_{\phi_d}(y_t\mid x,y_{<t})
\end{equation}
as the teacher disagreement. If conflicting teacher signals are a major cause of the multi-task
OPD failure, $c_t$ should be large for a substantial fraction of tokens.

\begin{figure}[t]
\centering
\includegraphics[width=\linewidth]{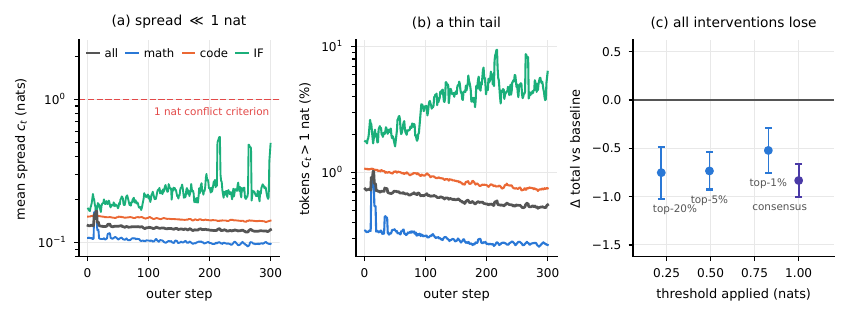}
\caption{\textbf{Teacher conflict is measurable but not the bottleneck.}
(a) Average teacher disagreement $c_t$ (log scale) and its distance to the 1-nat criterion. The legend applies to both panels (a) and (b).
(b) Fraction of tokens with $c_t > 1$, broken down by domain.
(c) Change in total score for four conflict interventions relative to the baseline. Blue points show the conflict mask, where the horizontal axis indicates the filtering threshold when removing top-$k\%$ tokens by $c_t$ quantile (larger $k$ means lower threshold). Purple points show the consensus method with a fixed threshold of 1 nat; tokens with $c_t \le 1$ use the average log-probs of all three teachers, while those with $c_t > 1$ keep the routed teacher.
Error bars denote validation standard error.}
\label{fig:conflict_zh}
\end{figure}

Figure~\ref{fig:conflict_zh} provides three tests of this hypothesis. The first two examine how
often strong disagreement occurs, while the third tests whether changing such tokens improves
training. First, disagreement is small and stable. We measured $c_t$ over the full training run. As we can see in Figure~\ref{fig:conflict_zh}a, the mean
value is 0.126 nat, and $c_t$ remains below 0.27 nat throughout the 300 training steps. Both
values are much smaller than the 1-nat conflict threshold (where the probability
ratio between the most and least likely teachers is about $e\approx2.7$). These results do not support widespread teacher conflict as the
main cause of the integration gap.

Second, high-conflict tokens are rare. Only 0.62\% of tokens satisfy $c_t>1$ on average. The rate
is 3.9\% on IF, where disagreement is largest, and 0.31\% on math
(Figure~\ref{fig:conflict_zh}b). A maximum value of 30.58 nat shows that extreme conflict can
occur, but it is limited to a small number of tokens.

Third, we tested whether directly changing high-conflict tokens improves training. We used two
interventions. A conflict mask removes the top 1\%, 5\%, or 20\% of tokens by $c_t$ from the
distillation loss, using thresholds of 0.83, 0.49, and 0.22 nat. We rescaled the remaining weights
to keep the total loss scale fixed. A consensus target replaces the hard-routed teacher target
with the average of the three teachers' log-probabilities when $c_t\le1$; it keeps the routed
teacher when $c_t>1$ (purple points in Figure~\ref{fig:conflict_zh}c).

Under the same settings, the three conflict masks reduce the total score by $0.52$, $0.73$, and
$0.75$ points, respectively, relative to the baseline. The consensus target reduces it by $0.83$
points. High-conflict tokens may be irrelevant noise, or they may carry domain-specific
information. Token-level disagreement alone cannot tell these cases apart. This may explain why
removing or replacing such tokens hurts performance.

\textbf{The teacher-conflict experiments show that token-level teacher disagreement under this metric is unlikely to be the dominant bottleneck.}
We therefore examine how much training signal each domain provides to the shared student and how
the available training resources are distributed across domains.

\subsection{Measuring training imbalance across domains}
\label{sec:diagnosis:measurement}

We study how much training signal each domain receives and how this amount changes during
training. We first measure the number of valid response tokens and the average per-token reward
for each domain. We then examine how these quantities change over the training trajectory. Finally,
we study the effect of using old student probabilities in later inner updates from the same rollout
batch. These measurements cover three sources of imbalance, namely token counts, reward magnitudes, and
updates based on old student probabilities.

\begin{figure}[t]
\centering
\includegraphics[width=\linewidth]{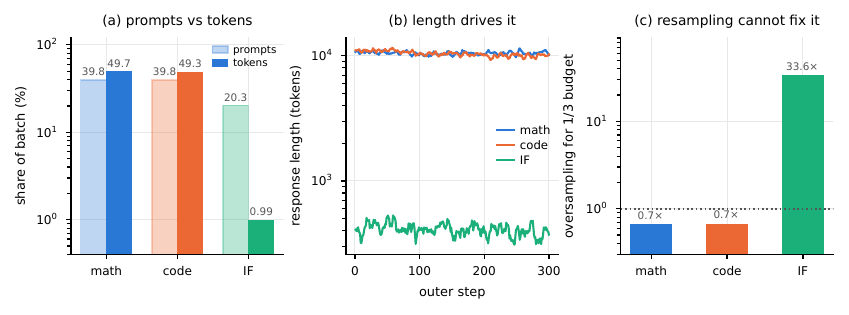}
\caption{\textbf{Prompt share and gradient-token share are decoupled.}
(a) Whole-run averages of prompt share versus token share per domain (log scale; light bars for prompts, solid bars for tokens). IF accounts for $\approx 20\%$ of prompts but only $\approx 1\%$ of gradient tokens.
(b) Average response length (log scale) accounts for this difference, with $\approx 10,500$ tokens for math/code versus $\approx 409$ tokens for IF.
(c) Balancing token share ($1/3$ per domain) solely via oversampling requires a $33.6\times$ prompt multiplier for IF. This severely reduces math and code prompts per batch, making long-chain supervision unsustainable.}
\label{fig:budget_zh}
\end{figure}

\paragraph{Token imbalance across domains.}
The training loss uses token-mean aggregation, so every valid response token contributes to the loss
once. The raw token share of domain $d$ is therefore
\begin{equation}
\label{eq:token_share_zh}
s_d^{\mathrm{tok}}
=\frac{n_dL_d}{\sum_{j\in\mathcal{D}}n_jL_j},
\end{equation}
where $n_d$ is the number of prompts of that domain and $L_d$ is the average response length.
Figure~\ref{fig:budget_zh} shows that the two shares are completely decoupled along the whole
trajectory, where the prompt share is fixed by the sampler at 39.8\%/39.8\%/20.3\%, whereas the token
share is 49.7\%/49.3\%/0.99\% (Figure~\ref{fig:budget_zh}a).
Across all 300 steps the token share of IF never exceeds 1.65\% and never falls below 0.44\%, so
this is not an incidental property of a particular batch.
The cause is given directly by length (Figure~\ref{fig:budget_zh}b): the average response length of
math and code is about 10{,}500 tokens while that of IF is only 409 tokens, a difference of more than
25 times; \Eqref{eq:token_share_zh} is therefore almost entirely determined by length.
This also shows that simply raising the sampling frequency of IF is not a viable fix. For IF to
obtain $1/3$ of the token budget, its number of prompts would have to be scaled up about 33.6 times
(Figure~\ref{fig:budget_zh}c), and the math and code prompts in a batch would be squeezed to the
point where long-chain supervision can no longer be maintained.

\begin{figure}[t]
\centering
\includegraphics[width=\linewidth]{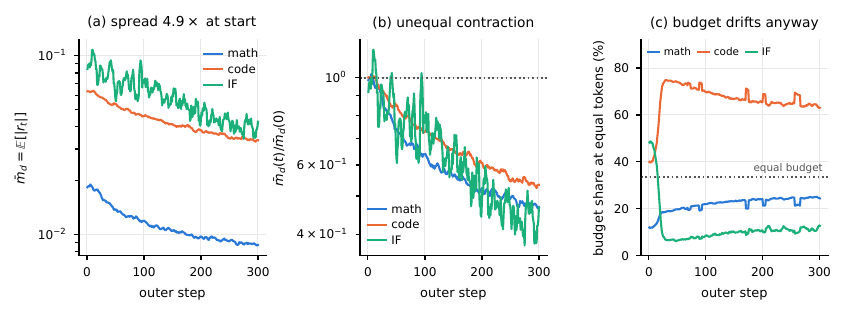}
\caption{\textbf{The effective budget drifts with distillation progress.}
(a) Per-token reward magnitude $\bar m_d$ of each domain (log axis), differing by 4.9$\times$ early on;
(b) after normalizing to their respective initial values, the three domains shrink at different
rates, which shows that $\bar m_d$ measures the remaining teacher--student gap;
(c) on a trajectory where $s_d^{\mathrm{tok}}$ is flattened to $1/3$,
\Eqref{eq:budget_proxy_zh} degenerates to
$B_d\propto\bar m_d$, so the curves show the budget drift caused by the gap alone. Within 25 steps
the share of IF falls below the dashed line to about 9\%, while code rises to 63.8\%. Flattening the
token share does not lock the budget in place.}
\label{fig:gapdyn_zh}
\end{figure}

\paragraph{Reward magnitudes also affect the update budget.}
In addition to the number of tokens, the strength of each update depends on the reward magnitude.
As a simple estimate of the update budget, the contribution of a domain is
\begin{equation}
\label{eq:budget_proxy_zh}
B_d\ \propto\ s_d^{\mathrm{tok}}\,\bar m_d,
\qquad
\bar m_d=\mathbb{E}_{t\in d}[\lvert r_t\rvert].
\end{equation}
Figure~\ref{fig:gapdyn_zh}a and \Eqref{eq:budget_proxy_zh} show how the reward
magnitude changes during training. Early in training, $\bar m_d$ is 0.019 for math, 0.063 for
code, and 0.091 for IF. The largest value is 4.9 times the smallest. Because $\bar m_d$ measures
the average difference between the student and teacher policies, it should decrease as the student
approaches its teacher during training. The decrease is different across domains. Along the same
trajectory, IF shrinks by 2.4 times, math by 2.1 times, and code by 1.9 times
(Figure~\ref{fig:gapdyn_zh}b).
Here $\bar m_d$ is the average per-token reward magnitude for domain $d$. It is the expectation
of $|r_t|$ over the tokens in that domain and measures the average effect of one token on the
parameter update.

The effect of the reward magnitude becomes clear after the token shares are balanced. When
$s_d^{\mathrm{tok}}$ is fixed at $1/3$ for every domain, the training contribution of each domain
depends only on $\bar m_d$ in \Eqref{eq:budget_proxy_zh}. Within 25 steps, the budget
share of IF drops from 48.7\% to around 9\% and ends at
11.4\%, while that of code rises from 39.6\% to 63.8\%
(Figure~\ref{fig:gapdyn_zh}c). Thus, balancing the token counts does not keep the training
contributions balanced in later steps because the reward magnitudes differ across domains.

\begin{figure}[t]
\centering
\includegraphics[width=0.72\linewidth]{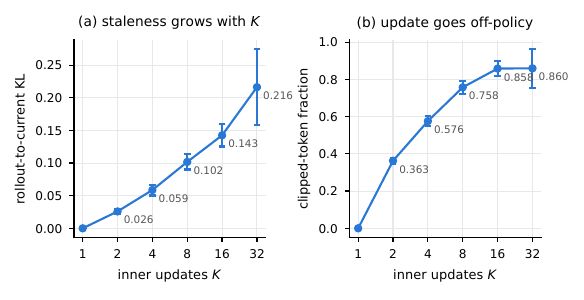}
\caption{\textbf{Repeated inner updates change the student policy within a rollout batch.} The
horizontal axis is the number of inner updates $K$ per rollout batch.
(a) Within the same batch, the KL between the rollout policy and the current student policy grows
monotonically with $K$;
(b) the fraction of tokens clipped by PPO rises with $K$. At $K{=}1$ both are 0, and the dense reward
is then consistent with the current student.}
\label{fig:stale_zh}
\end{figure}

\paragraph{Multiple inner updates make the reward stale.}
To reduce the cost of generation, practical training usually performs $K$ inner updates on one
large rollout batch. The teacher remains fixed, while the student changes after the first inner
update. If later updates still use the student probabilities computed during rollout,
$\log\pi_\theta(v\mid x,y_{<t})$ and $\tilde\pi_\theta(v\mid x,y_{<t})$ ($v\in\mathcal{S}_t$), then
the student-dependent part of \Eqref{eq:opd_reward_zh} is inconsistent with the current
policy.
Figure~\ref{fig:stale_zh} measures the policy shift by the KL between the rollout policy
and the current policy within the same batch, and its value rises monotonically with $K$, growing
from 0 at $K{=}1$ (no shift by definition) to 0.059 at $K{=}4$ and 0.216 at $K{=}32$
(Figure~\ref{fig:stale_zh}a); the fraction of tokens clipped by PPO rises with $K$, from 0 to 0.86
(Figure~\ref{fig:stale_zh}b).
Therefore, in a high-throughput setting, most tokens are updated after the student has already
drifted away from the rollout policy. The dense reward is still computed from the student and
teacher probabilities at rollout time.

These measurements identify three parts of the training signal that need to be controlled. The first is the token budget, followed by reward magnitude and reward freshness.
the token budget across domains, the teacher--student gap during training, and the reward delay
inside inner updates. The next section presents one method for each part.

\section{From Diagnosis to Method: Open-MOPD}
\label{sec:method_zh}

\begin{figure}[t]
\centering
\includegraphics[width=\linewidth]{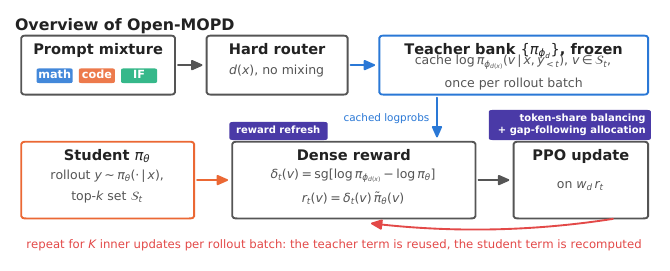}
\caption{\textbf{Overview of Open-MOPD.} 
Prompts are hard-routed by domain label. During rollout, the teacher computes and caches log-probabilities on $\mathcal{S}_t$ once. The student generates responses under its own distribution. Together, both terms form the dense reward for PPO updates. Within each rollout batch, the inner update repeats $K$ times, reusing the teacher term while recomputing the student term via reward refresh. 
Badges highlight where each mechanism acts: token-share balancing and gap-following allocation determine the domain loss weight $w_d$, while reward refresh rebuilds $r_t$ in every inner update.}
\label{fig:method_zh}
\end{figure}

Open-MOPD keeps the teacher routing and the on-policy distillation objective of
Section~\ref{sec:recipe:objective}, and changes only how the optimization budget is allocated and
how the reward components are computed.
The three mechanisms correspond to the three measurements of Section~\ref{sec:diagnosis:measurement}:
\emph{token-share balancing} controls the domain token budget within a batch,
\emph{gap-following allocation} adjusts the budget over the course of training according to the
remaining distillation gap,
\emph{reward refresh} refreshes the student-dependent reward across the several inner updates of one
rollout.
They act on different time scales, so they can be validated independently and can also be combined
into a single training recipe.
Figure~\ref{fig:method_zh} gives the complete method, showing the data flow of one training iteration, with
badges marking the stage that each mechanism rewrites.

\subsection{Token-Share Balancing}
\label{sec:method_zh:share}

Token-share balancing assigns a fixed weight to each domain's token-mean loss. Let $g_d^\star$ be the target domain budget, where $\sum_d g_d^\star=1$. In each batch, we compute $s_d^{\mathrm{tok}}$ from the attention mask and weight the loss of domain $d$ by
\begin{equation}
\label{eq:share}
w_d^{\mathrm{share}} = \frac{g_d^\star}{s_d^{\mathrm{tok}}}.
\end{equation}
After this weighting, the effective share of domain $d$ is
\begin{equation*}
\frac{w_d^{\mathrm{share}} s_d^{\mathrm{tok}}}{\sum_j w_j^{\mathrm{share}} s_j^{\mathrm{tok}}} = g_d^\star.
\end{equation*}
This directly controls the actual token ratio in the loss, without relying on stable response lengths or oversampling short answers.

The main recipe uses the equal-share target $g^\star=(1/3,1/3,1/3)$, which needs no corpus-specific
tuning. On the token shares shown in Figure~\ref{fig:budget_zh},
it yields weights of 0.69 for math, 0.66 for code and 32.7 for IF. Every IF token is amplified about
48 times to compensate for its 25-fold length disadvantage,
while \Eqref{eq:share} guarantees that the weighted shares of the three domains are
exactly 33.33\%. Compared with changing only the prompt sampling rates, token-share balancing
preserves the diversity of math and code prompts while giving the shorter IF responses a meaningful
contribution to training.

Token-share balancing changes how much each domain contributes to the loss through its token weight; it does not change the sampling frequency or the reward magnitude of individual tokens. We intentionally keep this design because the size of the reward shows the gap between the teacher and the student, and simply normalizing it would remove useful information about learning progress.

\subsection{Gap-Following Allocation}
\label{sec:method_zh:gap}
% TODO
Let $m_d$ be the running per-token reward magnitude for domain $d$ (for example, an exponential moving average of $\lvert r\rvert$), which serves as an observable proxy for the remaining teacher--student gap. 
A naive idea is to normalize the loss with $m_d^{-\alpha}$, giving larger weights to domains
with smaller rewards. This may seem reasonable early in training because a small reward can
indicate that a domain is learning slowly. However, $m_d$ also tracks the remaining distillation
gap: as a domain approaches its teacher, its $m_d$ becomes smaller. Inverse normalization therefore
assigns more budget to domains that have already made more progress.

This forms an unstable feedback loop. As a domain converges, its $m_d$ decreases, which in turn increases its assigned weight $m_d^{-\alpha}$. The domain then receives even more training budget, accelerating its convergence and shrinking $m_d$ further. Without any balancing force, the weights continuously diverge and eventually crash the training. We observe this pattern directly in experiments: over the first 75 steps, $m_d$ for IF decreases 35.3-fold. Setting $\alpha=0.5$ causes the IF weight to rise from 26.7 to 90.7, while the code weight drops from 0.44 to 0.27. Therefore, we believe that the inverse rule is unstable when $\alpha$ is positive and not too small.

Gap-following allocation keeps the direction of the gap, interprets it as ``capability not yet
distilled'', and allocates the budget to the domains that still have a larger gap.
\begin{equation}
\label{eq:gap_zh}
\tilde w_d
=w_d^{\mathrm{share}}\cdot
\operatorname{Clamp}\!\left(
\left(\frac{m_d}{m_{\mathrm{ref}}}\right)^\alpha,\,0.05,\,20
\right),
\qquad
w_d^{\mathrm{gap}}
=\frac{\tilde w_d}{\sum_j \tilde w_j s_j^{\mathrm{tok}}},
\end{equation}
Here $\tilde w_d$ is the unnormalized weight before the final normalization. It combines the
token-share weight $w_d^{\mathrm{share}}$ from the previous section with a clipped gap factor.
The reference value $m_{\mathrm{ref}}$ is the mean of $m_d$ across domains in the current batch.
Thus, $(m_d/m_{\mathrm{ref}})^\alpha$ measures the reward magnitude of domain $d$ relative to the
other domains. We clip this factor to $[0.05,20]$ so that a sudden change in one domain's reward
does not make its training weight too small or too large. The divisor
$\sum_j\tilde w_js_j^{\mathrm{tok}}$ is the token-share-weighted mean of the weights, so after
normalization $\sum_d w_d^{\mathrm{gap}}s_d^{\mathrm{tok}}=1$ and the total loss scale of a batch
stays unchanged. Consequently, when a domain approaches its teacher, its $m_d$ and its budget fall
together, while domains with a larger teacher--student gap receive more updates. This allocation rule
therefore follows the remaining gap rather than normalizing reward magnitudes; Section~\ref{sec:ablation:gap}
shows its benefit on top of token-share balancing and the collapse that occurs when $\alpha$ takes a
negative sign.

\subsection{Reward Refresh}
\label{sec:method_zh:refresh}

The first two mechanisms deal with the budget between domains; reward refresh deals with temporal
inconsistency inside one rollout.
Our design draws on an insight from AsyncOPD~\citep{kang2026asyncopd}: when the student changes,
the student-dependent part of a reverse-KL signal should be recomputed with the current student.
We apply this insight to a different source of staleness by refreshing the student-dependent reward
before each of the $K$ inner updates in our synchronous multi-teacher pipeline. The student is
updated $K$ times within one rollout batch, while the rollout
trajectories come from the student before these updates. This creates a mismatch in the reward,
which reward refresh corrects. Suppose that one rollout is used for $K$ update minibatches. The
teacher needs only one prefill to compute
$\log\pi_{\phi_{d(x)}}(v\mid x,y_{<t})$, while the student probabilities are recomputed after
each update.
In the actor forward of the $k$-th inner update, reward refresh recomputes the student-dependent term
on the same set $\mathcal{S}_t$.
\begin{equation}
\label{eq:refresh_zh}
\begin{aligned}
\delta_t^{(k)}(v)
&=
\mathrm{sg}\!\Big[
\log\pi_{\phi_{d(x)}}(v\mid x,y_{<t})
-
\log\pi_\theta^{(k)}(v\mid x,y_{<t})
\Big],
\\[2pt]
r_t^{(k)}(v)
&=
\delta_t^{(k)}(v)\cdot\tilde\pi_\theta^{(k)}(v\mid x,y_{<t}),
\qquad v\in\mathcal{S}_t,\quad k=0,\ldots,K-1.
\end{aligned}
\end{equation}
Here $\tilde\pi_\theta^{(k)}$ is computed on the current student log-probabilities following
\Eqref{eq:opd_reward_zh}.
It is then aggregated into $r_t^{(k)}$ by \Eqref{eq:opd_pos_zh} and placed in the advantage
slot of PPO. The update is subsequently constructed with the
$w_dr_t^{(k)}$ of the current domain, where $w_d=w_d^{\mathrm{gap}}$ when gap-following allocation is
enabled and $w_d=w_d^{\mathrm{share}}$ otherwise.
Reward refresh reuses the student forward pass already required by PPO and adds no teacher
forward. It therefore makes each update use the difference between the current student and the
teacher.

At $K=1$, Eqs.~(3), (4), and (10) reduce to the objective in
Section~\ref{sec:recipe:objective}; at $K>1$, they repair only the part of the reward that explicitly
depends on the student. The sampled student states still come from rollout time, so reward refresh
removes the staleness that can be removed without another prefill.

\subsection{The Complete Algorithm}
\label{sec:method_zh:algorithm}

Table~\ref{tab:mechanism_summary_zh} summarizes the three mechanisms and the quantities they modify.
Together, they change token weighting, domain weighting, and reward evaluation in the shared student.
\begin{table}[t]
\caption{\textbf{Summary of the three Open-MOPD mechanisms.} Each mechanism modifies a different part of the training computation.}
\label{tab:mechanism_summary_zh}
\small
\setlength{\tabcolsep}{0pt}
\begin{tabular}{@{}>{\raggedright\arraybackslash}p{0.22\linewidth}
                @{\hspace{0.5em}}
                >{\raggedright\arraybackslash}p{0.26\linewidth}
                @{\hspace{0.5em}}
                >{\raggedright\arraybackslash}p{\dimexpr\linewidth-0.48\linewidth-1em\relax}@{}}
\toprule
Mechanism & Measurement & Intervention \\
\midrule
token-share balancing & Response-token share $s_d^{\mathrm{tok}}$ in the current batch & Set $w_d^{\mathrm{share}}=g_d^\star/s_d^{\mathrm{tok}}$ to equalise the domains' token shares in the loss. \\
gap-following allocation & Running mean reward magnitude $m_d$ for each domain & Multiply the share weight by the clipped factor $(m_d/m_{\mathrm{ref}})^\alpha$, giving more weight to domains with a larger current reward signal. \\
reward refresh & Student log-probabilities at each inner update & Recompute the student-dependent reward while reusing the cached teacher log-probabilities. \\
\bottomrule
\end{tabular}
\end{table}

Algorithm~\ref{alg:openmopd_iteration} summarizes the complete training loop. The teacher
log-probabilities are computed once per rollout, while the student-dependent reward term is
recomputed before each inner update.

\begin{algorithm}[t]
\caption{Open-MOPD training loop}
\label{alg:openmopd_iteration}
\begin{algorithmic}[1]
\Require Domain-mixed sampler, student $\pi_\theta$, teachers $\{\pi_{\phi_d}\}$, inner-step count $K$
\Ensure Updated student $\pi_\theta$
\For{each training step $t$}
  \State Sample rollout batch $\mathcal{R}_t$ from $\pi_\theta$ and route prompts by domain label.
  \State Compute teacher log-probabilities $\ell_{\phi,d(x^{(b)})}^{(b)}$ for each sampled response.
  \State Compute response-token shares $s_d^{\mathrm{tok}}$ and weights $w_d^{\mathrm{share}}$ using \Eqref{eq:share}.
  \State Update reward means $m_d$ and compute gap weights $w_d$ using \Eqref{eq:gap_zh}.
  \For{each inner minibatch $\mathcal{M}_{t,k}$, $k=0,\ldots,K-1$}
    \State Recompute current-student log-probabilities and refresh $r_t^{(k)}$ using \Eqref{eq:refresh_zh}.
    \State Update $\theta$ with the PPO objective using $w_d r_t^{(k)}$.
  \EndFor
\EndFor
\end{algorithmic}
\end{algorithm}

Together, these three mechanisms address the problems found in our diagnosis. Token-share
balancing fixes the token budget, gap-following allocation changes each domain's share of updates
as training progresses, and reward refresh keeps the feedback up to date.

\section{Ablation Studies}
\label{sec:ablation}

Token-share balancing controls the response-token share across domains, gap-following allocation
controls how the update budget follows the remaining teacher--student gap, and reward refresh
recomputes the student-dependent reward before each inner update. We first verify these effects one
mechanism at a time and measure the resulting change in the corresponding domain score. We then
evaluate the combined recipe by its reduction of the integration gap. For reward refresh, we also
measure the runtime cost.

\begin{table}[t]
\caption{\textbf{From Naive M-OPD to the full recipe.} 
Each row changes one thing from the row above, as shown in the first column. 
The three middle columns show whether token-share balancing, gap-following allocation, and reward refresh are turned on. $K$ is the number of inner updates per rollout batch.}
\label{tab:ablation_zh}
\centering
\small
\setlength{\tabcolsep}{2.6pt}
\begin{tabular}{@{}l ccc c c c ccc cc@{}}
\toprule
& \multicolumn{3}{c}{Mechanism} && && \multicolumn{5}{c}{Six datasets} \\
\cmidrule(lr){2-4}\cmidrule(lr){8-12}
Configuration & share & gap & refresh && $K$ && Math & Code & IF & Total & $\Delta$ \\
\midrule
Naive M-OPD &  &  &  && 1 && 21.26 & 19.26 & 43.64 & 28.05 & --- \\
\quad +share & \checkmark &  &  && 1 && 20.55 & 19.57 & 47.53 & 29.22 & $+1.17$ \\
\quad +gap & \checkmark & \checkmark &  && 1 && 21.00 & 19.31 & 49.50 & 29.94 & $+1.89$ \\
\midrule
\multicolumn{12}{@{}l}{\textit{Switching to the $K{=}4$ throughput setting (256 prompts per update, $4\times$ rollout batch)}} \\
\routegrey{Naive M-OPD (same-setting control)} & \routegrey{} & \routegrey{} & \routegrey{}
  && \routegrey{4} && \routegrey{21.62} & \routegrey{19.72} & \routegrey{46.49}
  & \routegrey{29.28} & \routegrey{$+1.23$} \\
\quad +share+gap & \checkmark & \checkmark &  && 4 && 23.05 & 21.07 & 47.16 & 30.43 & $+2.38$ \\
Open-MOPD & \checkmark & \checkmark & \checkmark && 4 && 22.42 & 21.73 & 49.58 & \textbf{31.24} & $\mathbf{+3.19}$ \\
\bottomrule
\end{tabular}
\end{table}

\subsection{Token-Share Balancing}
\label{sec:ablation:share}

The first test checks the mechanism directly. With token-share balancing enabled (Section~\ref{sec:method_zh:share}), the weighted token
share $w_ds_d^{\mathrm{tok}}$ stays at $33.33\%$ over all 300 steps (marked by the dashed line in
Figure~\ref{fig:gapfollow_zh}a); the same sampler with it switched off hands $99\%$ of the gradient
tokens to math and code and leaves only about $1\%$ to IF (Figure~\ref{fig:budget_zh}a).
The first cause of imbalance measured in Section~\ref{sec:diagnosis:measurement} is therefore removed
entirely. In Table~\ref{tab:ablation_zh}, token-share balancing gives a gain of $+1.17$ points, almost
entirely from IF; math and code barely move. This shows that, under naive M-OPD, most token loss
is spent on math and code, where the student is already close to its teacher. Token-share
balancing increases the token share of IF, which receives very little training otherwise, and
improves the overall M-OPD score.

\subsection{Gap-Following Allocation}
\label{sec:ablation:gap}

\begin{figure}[t]
\centering
\includegraphics[width=\linewidth]{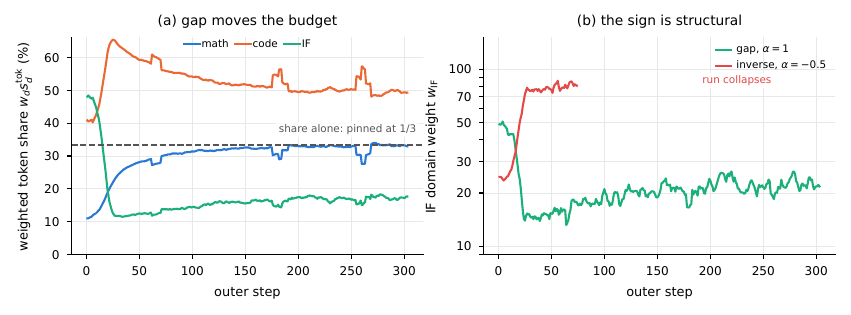}
\caption{
\textbf{Gap-following allocation dynamically changes the budget trajectory.}
(a) Weighted token share $w_d s_d^{\mathrm{tok}}$: static token-share balancing keeps the share fixed at $1/3$ (dashed line). Adding gap-following allocation ($\alpha{=}1$) dynamically adjusts the budget based on the remaining teacher--student gap. Domains that converge quickly lose budget, while those with larger gaps receive more resources (e.g., code exceeds $50\%$, whereas IF drops to $\sim 17\%$).
(b) IF domain weight (log scale): setting $\alpha{=}1$ properly reduces its weight, whereas the inverse formulation ($\alpha{=}-0.5$) creates an unstable feedback loop, causing training to collapse at step 74.
}
\label{fig:gapfollow_zh}
\end{figure}

We add gap-following allocation on top of token-share balancing. Unlike static balancing, which keeps
every domain at around 1/3, gap-following allocation changes domain ratios dynamically as training
goes on (Figure~\ref{fig:gapfollow_zh}a). In this setting, the budget goes to the domain with the
largest teacher–student gap $\bar{m}_d$. Averages calculated over the middle window of the baseline
trajectory show that $\bar{m}_d$ is $0.028$ for code, $0.008$ for math, and $0.003$ for IF
($2.16\times$, $0.61\times$, and $0.23\times$ the average, respectively). From the first 25 steps
to steps 200–300, the math share grows from 15.1\% to 32.6\%, the IF share drops from 34.5\% to 17.1\%,
and code gets the most budget (averaging 55.4\% and peaking at 87.6\%). When tested alone without
reward refresh, this mechanism brings a $+0.72$-point improvement ($K=1$ ladder, Table~\ref{tab:ablation_zh}).

Reversing the factor gives more budget to domains with smaller gaps. This favors domains that the
student already handles well and creates a positive feedback loop. In our experiment, the IF gap
shrinks by $32.5\times$, its weight rises from 24.4 to 80.9, and training stops at step 74
(Figure~\ref{fig:gapfollow_zh}b). The sign of $\alpha$ is therefore fixed by the direction of the
gap.

\subsection{Reward Refresh}
\label{sec:ablation:refresh}

Reward refresh (Section~\ref{sec:method_zh:refresh}) moves the student-dependent part of the dense
reward into every inner update. The teacher log-probs are still prefilled once at the top of the
rollout batch and reused throughout, and the student log-probabilities are read from the actor
forward that PPO already performs. Reward refresh therefore adds no extra student forward. Without
refresh, the dense reward uses the student probabilities saved at rollout time; with refresh, it uses
the student probabilities from the current inner update. The teacher term is computed once per outer
step in both cases; only the student parameters used to evaluate the reward change.

\begin{figure}[t]
\centering
\includegraphics[width=0.55\linewidth]{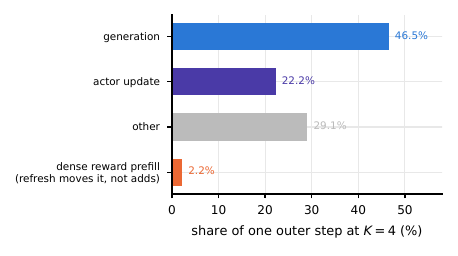}
\caption{
\textbf{Composition of one outer step at $K{=}4$.} Averaged over steady training steps.
The dense-reward computation accounts for $2.2\%$ of each step. Reward refresh changes only when the
student probabilities are read for this computation, without changing total throughput.
}
\label{fig:refresh_zh}
\end{figure}

We measure the extra runtime and performance gain from reward refresh. At $K{=}4$, the
dense-reward computation takes 27.3\,s with refresh and 27.8\,s without refresh. These values
account for $2.10\%$ and $2.12\%$ of one outer step, respectively (Figure~\ref{fig:refresh_zh}).
The full step takes 1298\,s with refresh and 1313\,s without it. The difference is small and
falls within the step-to-step variation, so \textbf{reward refresh adds no measurable extra runtime cost}.
On top of token-share balancing and gap-following allocation, reward refresh gives a further
$+0.81$ points at $K{=}4$ (Table~\ref{tab:ablation_zh}). This gain supports the effectiveness of
the mechanism and completes the Open-MOPD recipe.

\section{Related Work}
\label{sec:related}

\subsection{On-Policy Distillation and Multi-Teacher Extensions}
\label{sec:related:opd}

\paragraph{Distillation on a fixed corpus.}
Knowledge distillation trains a student to match the teacher's output distribution
\citep{hinton2015distilling}. For sequence models, the teacher usually generates responses in
advance, and the student learns from this fixed corpus
\citep{kim2016sequence}. The student therefore trains on prefixes produced by the teacher, while
inference uses prefixes produced by the student. This distribution shift creates the exposure-bias
problem in sequence prediction
\citep{ranzato2015sequence, bengio2015scheduled}. A large difference in teacher and student
capacity can also make distillation difficult
\citep{cho2019efficacy, mirzadeh2019improved, li2025small}.

\paragraph{On-policy distillation.}
On-policy distillation (OPD) trains the student on its own rollouts and uses teacher feedback on
the prefixes that the student actually visits \citep{gu2023minillm, agarwal2023policy}. Recent work
improves this process by changing the divergence, the sampling rule, or the token-level training
signal \citep{ko2024distillm, ko2025distillm, xu2024speculative, li2026rethinking}. Other studies
examine training stability and failure modes, including unreliable feedback on long or drifted
prefixes \citep{fu2026revisiting}. A recent survey organizes these methods by their feedback
signal, teacher access, and optimization rule \citep{song2026survey}. Most existing work still uses
one teacher. Our work studies how several teachers share the updates of one student.

\paragraph{Staleness under batch reuse.}
PPO supports multiple minibatch updates within one rollout cycle
\citep{schulman2017proximal}. In asynchronous RL, policy lag occurs when rollouts come from a policy
that is several updates behind the learner
\citep{fu2025areal, zhong2025streamrl, sheng2025laminar, yan2025areal, gao2025rollpacker,
sheng2024hybridflow}. Prior work studies how much stale data RL systems can tolerate
\citep{zheng2025prosperity, li20253po}, and recent work examines staleness in OPD
\citep{kang2026asyncopd, zheng2026blockwise}. In our synchronous pipeline, each rollout batch is
partitioned into $K$ minibatches and used for sequential student updates, so the student-dependent
reward can become stale as the student changes. Reward refresh recomputes this term before each
inner update using the student probabilities from PPO's existing actor forward, adding no extra
student forward.

\paragraph{From Single-Teacher to Multi-Teacher OPD.}
MOPD extends OPD to multiple domain teachers through a standard three-stage recipe: train
domain specialists from a shared SFT model, route each student rollout to its domain teacher,
and use token-level teacher feedback for training
\citep{xiao2026mimo, ma2026mopd}. This recipe has been used in several public models.
Nemotron-Cascade 2 distills strong intermediate teachers to recover capabilities lost during
Cascade RL, while Agents-A1 combines six domain teachers and normalizes the loss across
responses and domains \citep{yang2026nemotroncascade, bai2026agentsa1}. At a larger scale,
DeepSeek-V4 distills more than ten teachers, and Kimi K3 uses nine teachers defined by domain
and reasoning effort \citep{deepseekai2026deepseek, kimi2026k3}. These studies show that MOPD
can integrate specialists at different model scales. Open-MOPD studies how to balance the
training received by different domains during this integration.

\subsection{Integrating Domain Experts}
\label{sec:related:integration}

\paragraph{Training Domain Specialists.}
Reinforcement learning on a single domain can produce a strong specialist for that domain.
The success of DeepSeekMath and DeepSeek-R1 in mathematical reasoning provided a practical
recipe for LLM reinforcement learning
\citep{shao2024deepseekmath, deepseekai2025deepseek}. DAPO further developed this recipe for
large-scale training \citep{yu2025dapo}. Domain-specific RL has since expanded to software
engineering, search, and instruction following
\citep{jain2025r2e, wei2025swe, jin2025search, pyatkin2025generalizing}. Each pipeline produces
one specialist, while deployment usually requires one model that can handle all of these
domains. Our work studies how to integrate these specialists into one student through
multi-teacher OPD.

\paragraph{Integration in data space.}
A common way to combine domains is to train one model on mixed-domain data. Qwen3, for example,
uses general-domain RL to improve a wide range of tasks \citep{yang2025qwen3}. Another approach
trains the domains in sequence. Nemotron-Cascade applies a separate RL stage to each domain, so
each stage can use its own data and training settings \citep{wang2025nemotron}. In joint training,
the data mixture can strongly affect the final model. DoReMi studies this problem in pre-training,
while MoDoMoDo extends data-mixture optimization to multi-domain RLVR
\citep{xie2023doremi, liang2025modomodo}. Equal task sampling still does not guarantee equal
training: different tasks can produce gradients with very different magnitudes
\citep{wu2025imbalanced}. Open-MOPD studies how unequal response-token counts, different
teacher--student gaps, and outdated rewards create imbalance across domains in multi-teacher OPD.

\paragraph{Integration in weight space and in module space.}
A second route combines domain experts after they have been trained. Weight-space methods merge
task-specific parameter changes into one checkpoint. Task Arithmetic introduced this operation,
TIES-Merging resolves conflicting parameter signs, and DARE sparsifies model changes before they
are merged
\citep{ilharco2022editing, yadav2023ties, yu2023language}. Other methods retain the expert
structure during integration. BTM trains experts on different domains and combines them through
ensembling or parameter averaging, BTX turns their feed-forward layers into a routed mixture of
experts, and BTS connects frozen experts with lightweight stitch layers
\citep{li2022branch, sukhbaatar2024branch, zhang2025bts}. These methods integrate specialists
after separate training. Multi-teacher OPD trains one shared student from their outputs, and
Open-MOPD studies how to balance the updates received by different domains during this shared
training.

\section{Conclusion}
\label{sec:conclusion}

We build from scratch a fully open multi-teacher on-policy distillation pipeline---mixed-domain SFT,
three domain RL teachers, multi-teacher OPD---and use it to answer one concrete question, namely when
routing is already error-free, what prevents the capabilities of three experts from being written
into the same set of parameters at once.
Our experiments identify the allocation of the optimization budget as the main bottleneck. We identify
three separable contributors to imbalance: unequal token counts, unequal reward magnitudes, and outdated
rewards during repeated inner updates.
The three mechanisms of Open-MOPD---\textbf{token-share balancing, gap-following allocation, and
reward refresh}---correspond one by one to these three sources and are validated separately in ablation
experiments. Together, they reduce the integration gap from 3.50 points to 0.31 points, while the
recovery rate relative to RouteRL rises from $\mathbf{35.6\%}$ to $\mathbf{83.4\%}$.
We release the complete recipe and the mechanism implementations to support reproducible follow-up
work.

\clearpage
\bibliographystyle{unsrtnat}
\bibliography{references}

@misc{hinton2015distilling,
  title={{Distilling the Knowledge in a Neural Network}},
  author={Hinton, Geoffrey and Vinyals, Oriol and Dean, Jeff},
  year={2015},
  eprint={1503.02531},
  archivePrefix={arXiv}
}

@misc{bengio2015scheduled,
  title={{Scheduled Sampling for Sequence Prediction with Recurrent Neural Networks}},
  author={Bengio, Samy and Vinyals, Oriol and Jaitly, Navdeep and Shazeer, Noam},
  year={2015},
  eprint={1506.03099},
  archivePrefix={arXiv}
}

@misc{ranzato2015sequence,
  title={{Sequence Level Training with Recurrent Neural Networks}},
  author={Ranzato, Marc'Aurelio and Chopra, Sumit and Auli, Michael and Zaremba, Wojciech},
  year={2015},
  eprint={1511.06732},
  archivePrefix={arXiv}
}

@misc{kim2016sequence,
  title={{Sequence-Level Knowledge Distillation}},
  author={Kim, Yoon and Rush, Alexander M.},
  year={2016},
  eprint={1606.07947},
  archivePrefix={arXiv}
}

@misc{schulman2017proximal,
  title={{Proximal Policy Optimization Algorithms}},
  author={Schulman, John and Wolski, Filip and Dhariwal, Prafulla and Radford, Alec and Klimov, Oleg},
  year={2017},
  eprint={1707.06347},
  archivePrefix={arXiv}
}

@misc{mirzadeh2019improved,
  title={{Improved Knowledge Distillation via Teacher Assistant}},
  author={Mirzadeh, Seyed-Iman and Farajtabar, Mehrdad and Li, Ang and Levine, Nir and Matsukawa, Akihiro and Ghasemzadeh, Hassan},
  year={2019},
  eprint={1902.03393},
  archivePrefix={arXiv}
}

@misc{cho2019efficacy,
  title={{On the Efficacy of Knowledge Distillation}},
  author={Cho, Jang Hyun and Hariharan, Bharath},
  year={2019},
  eprint={1910.01348},
  archivePrefix={arXiv}
}

@misc{li2022branch,
  title={{Branch-Train-Merge: Embarrassingly Parallel Training of Expert Language Models}},
  author={Li, Margaret and Gururangan, Suchin and Dettmers, Tim and Lewis, Mike and Althoff, Tim and Smith, Noah A. and Zettlemoyer, Luke},
  year={2022},
  eprint={2208.03306},
  archivePrefix={arXiv}
}

@misc{ilharco2022editing,
  title={{Editing Models with Task Arithmetic}},
  author={Ilharco, Gabriel and Ribeiro, Marco Tulio and Wortsman, Mitchell and Gururangan, Suchin and Schmidt, Ludwig and Hajishirzi, Hannaneh and Farhadi, Ali},
  year={2022},
  eprint={2212.04089},
  archivePrefix={arXiv}
}

@misc{xie2023doremi,
  title={{DoReMi: Optimizing Data Mixtures Speeds Up Language Model Pretraining}},
  author={Xie, Sang Michael and Pham, Hieu and Dong, Xuanyi and Du, Nan and Liu, Hanxiao and Lu, Yifeng and Liang, Percy and Le, Quoc V. and Ma, Tengyu and Yu, Adams Wei},
  year={2023},
  eprint={2305.10429},
  archivePrefix={arXiv}
}

@misc{yu2023language,
  title={{Language Models are Super Mario: Absorbing Abilities from Homologous Models as a Free Lunch}},
  author={Yu, Le and Yu, Bowen and Yu, Haiyang and Huang, Fei and Li, Yongbin},
  year={2023},
  eprint={2311.03099},
  archivePrefix={arXiv}
}

@misc{gu2023minillm,
  title={{MiniLLM: On-Policy Distillation of Large Language Models}},
  author={Gu, Yuxian and Dong, Li and Wei, Furu and Huang, Minlie},
  year={2023},
  eprint={2306.08543},
  archivePrefix={arXiv}
}

@misc{agarwal2023policy,
  title={{On-Policy Distillation of Language Models: Learning from Self-Generated Mistakes}},
  author={Agarwal, Rishabh and Vieillard, Nino and Zhou, Yongchao and Stanczyk, Piotr and Ramos, Sabela and Geist, Matthieu and Bachem, Olivier},
  year={2023},
  eprint={2306.13649},
  archivePrefix={arXiv}
}

@misc{yadav2023ties,
  title={{TIES-Merging: Resolving Interference When Merging Models}},
  author={Yadav, Prateek and Tam, Derek and Choshen, Leshem and Raffel, Colin and Bansal, Mohit},
  year={2023},
  eprint={2306.01708},
  archivePrefix={arXiv}
}

@misc{sukhbaatar2024branch,
  title={{Branch-Train-MiX: Mixing Expert LLMs into a Mixture-of-Experts LLM}},
  author={Sukhbaatar, Sainbayar and Golovneva, Olga and Sharma, Vasu and Xu, Hu and Lin, Xi Victoria and Rozière, Baptiste and Kahn, Jacob and Li, Daniel and others},
  year={2024},
  eprint={2403.07816},
  archivePrefix={arXiv}
}

@misc{shao2024deepseekmath,
  title={{DeepSeekMath: Pushing the Limits of Mathematical Reasoning in Open Language Models}},
  author={Shao, Zhihong and Wang, Peiyi and Zhu, Qihao and Xu, Runxin and Song, Junxiao and Bi, Xiao and Zhang, Haowei and Zhang, Mingchuan and others},
  year={2024},
  eprint={2402.03300},
  archivePrefix={arXiv}
}

@misc{ko2024distillm,
  title={{DistiLLM: Towards Streamlined Distillation for Large Language Models}},
  author={Ko, Jongwoo and Kim, Sungnyun and Chen, Tianyi and Yun, Se-Young},
  year={2024},
  eprint={2402.03898},
  archivePrefix={arXiv}
}

@misc{sheng2024hybridflow,
  title={{HybridFlow: A Flexible and Efficient RLHF Framework}},
  author={Sheng, Guangming and Zhang, Chi and Ye, Zilingfeng and Wu, Xibin and Zhang, Wang and Zhang, Ru and Peng, Yanghua and Lin, Haibin and Wu, Chuan},
  year={2024},
  eprint={2409.19256},
  archivePrefix={arXiv}
}

@misc{xu2024speculative,
  title={{Speculative Knowledge Distillation: Bridging the Teacher-Student Gap Through Interleaved Sampling}},
  author={Xu, Wenda and Han, Rujun and Wang, Zifeng and Le, Long T. and Madeka, Dhruv and Li, Lei and Wang, William Yang and Agarwal, Rishabh and Lee, Chen-Yu and Pfister, Tomas},
  year={2024},
  eprint={2410.11325},
  archivePrefix={arXiv}
}

@misc{zhang2025bts,
  title={{BTS: Harmonizing Specialized Experts into a Generalist LLM}},
  author={Zhang, Qizhen and Bhargava, Prajjwal and Bi, Chloe and Cai, Chris X. and Foerster, Jakob and Fu, Jeremy and Koura, Punit Singh and Silva, Ruan and Shen, Sheng and Dinan, Emily and Gururangan, Suchin and Lewis, Mike},
  year={2025},
  eprint={2502.00075},
  archivePrefix={arXiv}
}

@misc{li20253po,
  title={{A-3PO: Accelerating Asynchronous LLM Training with Staleness-aware Proximal Policy Approximation}},
  author={Li, Xiaocan and Wu, Shiliang and Shen, Zheng},
  year={2025},
  eprint={2512.06547},
  archivePrefix={arXiv}
}

@misc{yan2025areal,
  title={{AReaL-Hex: Accommodating Asynchronous RL Training over Heterogeneous GPUs}},
  author={Yan, Ran and Jiang, Youhe and Wu, Tianyuan and Gao, Jiaxuan and Mei, Zhiyu and Fu, Wei and Mai, Haohui and Wang, Wei and Wu, Yi and Yuan, Binhang},
  year={2025},
  eprint={2511.00796},
  archivePrefix={arXiv}
}

@misc{fu2025areal,
  title={{AReaL: A Large-Scale Asynchronous Reinforcement Learning System for Language Reasoning}},
  author={Fu, Wei and Gao, Jiaxuan and Shen, Xujie and Zhu, Chen and Mei, Zhiyu and He, Chuyi and Xu, Shusheng and Wei, Guo and others},
  year={2025},
  eprint={2505.24298},
  archivePrefix={arXiv}
}

@misc{yu2025dapo,
  title={{DAPO: An Open-Source LLM Reinforcement Learning System at Scale}},
  author={Yu, Qiying and Zhang, Zheng and Zhu, Ruofei and Yuan, Yufeng and Zuo, Xiaochen and Yue, Yu and Dai, Weinan and Fan, Tiantian and others},
  year={2025},
  eprint={2503.14476},
  archivePrefix={arXiv}
}

@misc{deepseekai2025deepseek,
  title={{DeepSeek-R1: Incentivizing Reasoning Capability in LLMs via Reinforcement Learning}},
  author={DeepSeek-AI and Guo, Daya and Yang, Dejian and Zhang, Haowei and Song, Junxiao and Wang, Peiyi and Zhu, Qihao and Xu, Runxin and others},
  year={2025},
  eprint={2501.12948},
  archivePrefix={arXiv}
}

@misc{ko2025distillm,
  title={{DistiLLM-2: A Contrastive Approach Boosts the Distillation of LLMs}},
  author={Ko, Jongwoo and Chen, Tianyi and Kim, Sungnyun and Ding, Tianyu and Liang, Luming and Zharkov, Ilya and Yun, Se-Young},
  year={2025},
  eprint={2503.07067},
  archivePrefix={arXiv}
}

@misc{pyatkin2025generalizing,
  title={{Generalizing Verifiable Instruction Following}},
  author={Pyatkin, Valentina and Malik, Saumya and Graf, Victoria and Ivison, Hamish and Huang, Shengyi and Dasigi, Pradeep and Lambert, Nathan and Hajishirzi, Hannaneh},
  year={2025},
  eprint={2507.02833},
  archivePrefix={arXiv}
}

@misc{kimi2026k3,
  title={{Kimi K3: Open Frontier Intelligence}},
  author={{Kimi Team}},
  year={2026},
  eprint={2607.24653},
  archivePrefix={arXiv}
}

@misc{sheng2025laminar,
  title={{Laminar: A Scalable Asynchronous RL Post-Training Framework}},
  author={Sheng, Guangming and Tong, Yuxuan and Wan, Borui and Zhang, Wang and Jia, Chaobo and Wu, Xibin and Wu, Yuqi and Li, Xiang and others},
  year={2025},
  eprint={2510.12633},
  archivePrefix={arXiv}
}

@misc{bercovich2025llama,
  title={{Llama-Nemotron: Efficient Reasoning Models}},
  author={Bercovich, Akhiad and Levy, Itay and Golan, Izik and Dabbah, Mohammad and El-Yaniv, Ran and Puny, Omri and Galil, Ido and Moshe, Zach and others},
  year={2025},
  eprint={2505.00949},
  archivePrefix={arXiv}
}

@misc{wang2025nemotron,
  title={{Nemotron-Cascade: Scaling Cascaded Reinforcement Learning for General-Purpose Reasoning Models}},
  author={Wang, Boxin and Lee, Chankyu and Lee, Nayeon and Lin, Sheng-Chieh and Dai, Wenliang and Chen, Yang and Chen, Yangyi and Yang, Zhuolin and others},
  year={2025},
  eprint={2512.13607},
  archivePrefix={arXiv}
}

@misc{zheng2025prosperity,
  title={{Prosperity before Collapse: How Far Can Off-Policy RL Reach with Stale Data on LLMs?}},
  author={Zheng, Haizhong and Zhao, Jiawei and Chen, Beidi},
  year={2025},
  eprint={2510.01161},
  archivePrefix={arXiv}
}

@misc{yang2025qwen3,
  title={{Qwen3 Technical Report}},
  author={Yang, An and Li, Anfeng and Yang, Baosong and Zhang, Beichen and Hui, Binyuan and Zheng, Bo and Yu, Bowen and Gao, Chang and others},
  year={2025},
  eprint={2505.09388},
  archivePrefix={arXiv}
}

@misc{jain2025r2e,
  title={{R2E-Gym: Procedural Environments and Hybrid Verifiers for Scaling Open-Weights SWE Agents}},
  author={Jain, Naman and Singh, Jaskirat and Shetty, Manish and Zheng, Liang and Sen, Koushik and Stoica, Ion},
  year={2025},
  eprint={2504.07164},
  archivePrefix={arXiv}
}

@misc{gao2025rollpacker,
  title={{RollPacker: Mitigating Long-Tail Rollouts for Fast, Synchronous RL Post-Training}},
  author={Gao, Wei and Zhao, Yuheng and An, Dakai and Wu, Tianyuan and Cao, Lunxi and Xiong, Shaopan and Huang, Ju and Wang, Weixun and others},
  year={2025},
  eprint={2509.21009},
  archivePrefix={arXiv}
}

@misc{wei2025swe,
  title={{SWE-RL: Advancing LLM Reasoning via Reinforcement Learning on Open Software Evolution}},
  author={Wei, Yuxiang and Duchenne, Olivier and Copet, Jade and Carbonneaux, Quentin and Zhang, Lingming and Fried, Daniel and Synnaeve, Gabriel and Singh, Rishabh and Wang, Sida I.},
  year={2025},
  eprint={2502.18449},
  archivePrefix={arXiv}
}

@misc{jin2025search,
  title={{Search-R1: Training LLMs to Reason and Leverage Search Engines with Reinforcement Learning}},
  author={Jin, Bowen and Zeng, Hansi and Yue, Zhenrui and Yoon, Jinsung and Arik, Sercan and Wang, Dong and Zamani, Hamed and Han, Jiawei},
  year={2025},
  eprint={2503.09516},
  archivePrefix={arXiv}
}

@misc{li2025small,
  title={{Small Models Struggle to Learn from Strong Reasoners}},
  author={Li, Yuetai and Yue, Xiang and Xu, Zhangchen and Jiang, Fengqing and Niu, Luyao and Lin, Bill Yuchen and Ramasubramanian, Bhaskar and Poovendran, Radha},
  year={2025},
  eprint={2502.12143},
  archivePrefix={arXiv}
}

@misc{zhong2025streamrl,
  title={{StreamRL: Scalable, Heterogeneous, and Elastic RL for LLMs with Disaggregated Stream Generation}},
  author={Zhong, Yinmin and Zhang, Zili and Song, Xiaoniu and Hu, Hanpeng and Jin, Chao and Wu, Bingyang and Chen, Nuo and Chen, Yukun and others},
  year={2025},
  eprint={2504.15930},
  archivePrefix={arXiv}
}

@misc{song2026survey,
  title={{A Survey of On-Policy Distillation for Large Language Models}},
  author={Song, Mingyang and Zheng, Mao},
  year={2026},
  eprint={2604.00626},
  archivePrefix={arXiv}
}

@misc{kang2026asyncopd,
  title={{AsyncOPD: How Stale Can On-Policy Distillation Be?}},
  author={Kang, Wonjun and Galim, Kevin and Oh, Seunghyuk and Kang, Minjun and Park, Sanghyun and Kim, Donghoon and Lee, Minjae and Kim, Minseo and others},
  year={2026},
  eprint={2606.24143},
  archivePrefix={arXiv}
}

@misc{zheng2026blockwise,
  title={{Blockwise Policy-Drift Gating for On-Policy Distillation}},
  author={Zheng, Liwen and Jiang, Haiyun},
  year={2026},
  eprint={2606.24084},
  archivePrefix={arXiv}
}

@misc{deepseekai2026deepseek,
  title={{DeepSeek-V4: Towards Highly Efficient Million-Token Context Intelligence}},
  author={DeepSeek-AI and Xu, Anyi and Lin, Bangcai and Xue, Bing and Wang, Bingxuan and Xu, Bingzheng and Wu, Bochao and Zhang, Bowei and others},
  year={2026},
  eprint={2606.19348},
  archivePrefix={arXiv}
}

@misc{ma2026mopd,
  title={{MOPD: Multi-Teacher On-Policy Distillation for Capability Integration in LLM Post-Training}},
  author={Ma, Wenhan and Wei, Jianyu and Zhao, Liang and Zhang, Hailin and Xiao, Bangjun and Li, Lei and Yang, Qibin and Gao, Bofei and others},
  year={2026},
  eprint={2606.30406},
  archivePrefix={arXiv}
}

@misc{xiao2026mimo,
  title={{MiMo-V2-Flash Technical Report}},
  author={Core Team and Xiao, Bangjun and Xia, Bingquan and Yang, Bo and Gao, Bofei and Shen, Bowen and Zhang, Chen and He, Chenhong and others},
  year={2026},
  eprint={2601.02780},
  archivePrefix={arXiv}
}

@misc{li2026rethinking,
  title={{Rethinking On-Policy Distillation of Large Language Models: Phenomenology, Mechanism, and Recipe}},
  author={Li, Yaxuan and Zuo, Yuxin and He, Bingxiang and Zhang, Jinqian and Xiao, Chaojun and Qian, Cheng and Yu, Tianyu and Gao, Huan-ang and others},
  year={2026},
  eprint={2604.13016},
  archivePrefix={arXiv}
}

@misc{fu2026revisiting,
  title={{Revisiting On-Policy Distillation: Empirical Failure Modes and Simple Fixes}},
  author={Fu, Yuqian and Huang, Haohuan and Jiang, Kaiwen and Liu, Jiacai and Jiang, Zhuo and Zhu, Yuanheng and Zhao, Dongbin},
  year={2026},
  eprint={2603.25562},
  archivePrefix={arXiv}
}

@misc{smollm3,
  title={{SmolLM3}: Smol, Multilingual, Long-Context Reasoner},
  author={{Hugging Face}},
  year={2025},
  howpublished={\url{https://huggingface.co/blog/smollm3}}
}

@misc{yang2026nemotroncascade,
  title={{Nemotron-Cascade 2: Post-Training LLMs with Cascade RL and Multi-Domain On-Policy Distillation}},
  author={Yang, Zhuolin and Liu, Zihan and Chen, Yang and Dai, Wenliang and Wang, Boxin and Lin, Sheng-Chieh and Lee, Chankyu and Chen, Yangyi and Jiang, Dongfu and He, Jiafan and Pi, Renjie and Lam, Grace and Lee, Nayeon and Bukharin, Alexander and Shoeybi, Mohammad and Catanzaro, Bryan and Ping, Wei},
  year={2026},
  eprint={2603.19220},
  archivePrefix={arXiv}
}

@misc{bai2026agentsa1,
  title={{Scaling the Horizon, Not the Parameters: Reaching Trillion-Parameter Performance with a 35B Agent}},
  author={Bai, Lei and others},
  year={2026},
  eprint={2606.30616},
  archivePrefix={arXiv}
}

@misc{liang2025modomodo,
  title={{MoDoMoDo: Multi-Domain Data Mixtures for Multimodal LLM Reinforcement Learning}},
  author={Liang, Yiqing and Qiu, Jielin and Ding, Wenhao and Liu, Zuxin and Tompkin, James and Xu, Mengdi and Xia, Mengzhou and Tu, Zhengzhong and Shi, Laixi and Zhu, Jiacheng},
  year={2025},
  eprint={2505.24871},
  archivePrefix={arXiv}
}

@misc{wu2025imbalanced,
  title={{Imbalanced Gradients in RL Post-Training of Multi-Task LLMs}},
  author={Wu, Runzhe and Samanta, Ankur and Jain, Ayush and Fujimoto, Scott and Kwon, Jeongyeol and Kretzu, Ben and Yu, Youliang and Hassani, Kaveh and Vidolov, Boris and Efroni, Yonathan},
  year={2025},
  eprint={2510.19178},
  archivePrefix={arXiv}
}

\clearpage
\beginappendix
\section{Constructing the Open-MOPD Recipe}
\label{app:recipe}

In this section, we introduce how we obtained the final Open-MOPD recipe.
We first screen on math SFT for a base model that stably produces reasoning trajectories, and then
extend the selected base model to a mixed-domain SFT over math, code and instruction following. 
% This
% section records that construction process, so that the model choice in Section~\ref{sec:recipe:model}
% can be traced back to concrete experiments rather than being decided by parameter-count experience.

\subsection{Base Model Choice}
\label{app:recipe:base_model}

\paragraph{OpenR1 math comparison between 1.7B and 7B.}
Early screening applied the same 93,733 OpenR1 math examples to Qwen3-1.7B-Base and
Qwen2.5-7B-Base separately. Both runs use a 32,768-token SFT length limit, a global batch of 128 and
a learning rate of $4\times10^{-5}$, with 732 steps per epoch. 
Table~\ref{tab:app_openr1_base_curve} reports the complete curve over the first five epochs.

The problem with Qwen3-1.7B-Base is not a lack of generation length: under the 31K budget,
69.17--80.42\% of the samples still exhaust the limit at all five checkpoints, and only
20.00--31.25\% of the answers close \texttt{</think>}; adding epochs does not produce a monotone
improvement either, and AIME24 stays below 7.1\%. This means that a large fraction of the
trajectories seen by a subsequent RL or OPD stage contain only a reasoning prefix and never reach a
final answer. A drop in task score can then not be attributed uniquely to the reward, the routing or
cross-domain interference, because the base model and the data mixture themselves have not yet learnt
to end a reasoning trajectory reliably.

The same OpenR1 data shows the opposite trend on Qwen2.5-7B-Base: AIME24 rises from 16.25\% to
31.25\% and the truncation rate falls from 42.92\% to 12.92\%. This rules out the explanation that
``OpenR1 trajectories are inherently too long, so any base model would truncate'', and it shows that
enlarging the capacity of the base model does recover trajectory closure. Qwen2.5-7B-Base therefore
satisfies the capability condition, but it puts the student, the three RL teachers and every
mechanism ablation of the full recipe at 7B scale; for an open research baseline that has to be run
repeatedly, this cost is too high.

\paragraph{Choosing SmolLM3-3B-Base.}
The final recipe selects SmolLM3-3B-Base as the compromise between the two ends. Four epochs of
mixed-domain SFT over the three domains give $\pi_{\mathrm{mixsft}}$: OpenR1-Math-93k, the OCR-50k
subset sampled from the full OpenCodeReasoning set, and Instruction-Nemotron contribute roughly
37.3\%, 28.1\% and 34.6\% of the training response tokens respectively. The per-domain sample counts
are balanced by response token count, so that the 820K short IF answers do not drown out the fewer
but longer math and code trajectories.

Table~\ref{tab:app_base_choice_summary} summarises the selection evidence for the three candidate
base models. 
From the table, we can see that 3B is therefore not an arbitrary choice of a ``medium model'' but an operating point bracketed by two
failure boundaries: going down to 1.7B, generation is dominated by unclosed reasoning and truncation,
and mechanism attribution is no longer unique; going up to 7B, the single model is stronger, but the
end-to-end multi-teacher recipe and its repeated ablations exceed the resource boundary we set for a
community baseline. SmolLM3-3B-Base retains enough trajectory capacity while still allowing the final
Open-MOPD training to run on a single 8$\times$A100-80GB node.

\begin{table}[h]
\caption{Base-model choice for Open-MOPD. The table summarises directly the experimental results
used when constructing the recipe. The truncation evidence is reported in a dedicated row
below each base model.}
\vspace{5pt}
\label{tab:app_base_choice_summary}
\centering
\small
\setlength{\tabcolsep}{4pt}
\begin{tabular}{p{0.20\linewidth}p{0.25\linewidth}p{0.45\linewidth}}
\toprule
Base model & SFT data & Performance \\
\midrule

Qwen3-1.7B-Base
& OpenR1-Math-93k
& AIME24 best 7.08 \\
\multicolumn{3}{p{\dimexpr\linewidth-2\tabcolsep}}
{\hspace{0.5em}\footnotesize\textcolor{gray}{\textit{Truncation evidence:}
69.17--80.42\%; capacity and trajectory closure insufficient.}}
\\[3pt]

SmolLM3-3B-Base
& Three-domain mixSFT
& AIME24 15.63 \\
\multicolumn{3}{p{\dimexpr\linewidth-2\tabcolsep}}
{\hspace{0.5em}\footnotesize\textcolor{gray}{\textit{Truncation evidence:}
15.2\%; chosen as the base model of the recipe.}}
\\[3pt]

Qwen2.5-7B-Base
& OpenR1-Math-93k
& AIME24 31.25 \\
\multicolumn{3}{p{\dimexpr\linewidth-2\tabcolsep}}
{\hspace{0.5em}\footnotesize\textcolor{gray}{\textit{Truncation evidence:}
12.92\%; capable enough, but a full multi-teacher ablation is too costly.}}
\\

\bottomrule
\end{tabular}
\end{table}

\begin{table}[t]
\caption{Base-model screening under the same OpenR1-Math-93k SFT. Each cell is
AIME24 accuracy / generation truncation rate (\%). Qwen3-1.7B-Base mainly generates long, unclosed
reasoning trajectories at every observed checkpoint; Qwen2.5-7B-Base instead raises accuracy and
lowers truncation together as SFT proceeds.}
\label{tab:app_openr1_base_curve}
\centering
\small
\setlength{\tabcolsep}{5pt}
\begin{tabular}{c c c c}
\toprule
Epoch & Step & Qwen3-1.7B-Base & Qwen2.5-7B-Base \\
\midrule
1 & 732  & 4.58 / 75.42 & 16.25 / 42.92 \\
2 & 1464 & \textbf{7.08} / 72.50 & 20.83 / 35.42 \\
3 & 2196 & 6.67 / 80.42 & 30.83 / 22.08 \\
4 & 2928 & 5.83 / \textbf{69.17} & 30.42 / 19.58 \\
5 & 3660 & 6.67 / 69.58 & \textbf{31.25} / \textbf{12.92} \\
\bottomrule
\end{tabular}
\end{table}

\subsection{Stage Hyperparameters}
\label{app:recipe:hparams}

This section lists all hyperparameters of the four stages of the recipe, taken from the training
configurations that were actually run rather than from recommended values compiled after the fact.
Table~\ref{tab:app_hp_sft_rl_zh} gives the mixed-domain SFT and the three domain RL teachers, and
Table~\ref{tab:app_hp_opd_zh} gives single-domain OPD (the RouteOPD oracle of the main text) and
multi-teacher OPD. Items not listed use the verl defaults.

\paragraph{Mixed-domain SFT and domain RL teachers.}
The three teachers share one GRPO configuration and differ only in data, sequence length, rollout
group size and number of steps: IF answers are short, so it uses a smaller response limit, smaller
groups and more steps; math and code use a 30,000-token response limit to accommodate the full
reasoning chain. All three disable the KL penalty and apply group filtering on accuracy (the samples
of one prompt are discarded if they are all correct or all wrong, with at most 8 generation batches
resampled, which the dynamic sampling trick proposed by DAPO), so that the gradient comes only from groups that discriminate.

\begin{table}[t]
\caption{Hyperparameters of stages one and two. SFT is standard supervised fine-tuning; the three RL
teachers are all GRPO, each initialised from $\pi_{\mathrm{mixsft}}$ and trained only on the
verifiable reward of its own domain. World size is given as number of nodes $\times$
GPUs per node, all A100-80GB. ``---'' means the item does not apply to that stage.}
\label{tab:app_hp_sft_rl_zh}
\centering
\small
\setlength{\tabcolsep}{4.5pt}
\begin{tabular}{@{}l cccc@{}}
\toprule
& Mixed-domain SFT & RL: math & RL: code & RL: IF \\
\midrule
World size & 8$\times$8 & 8$\times$8 & 8$\times$8 & 4$\times$8 \\
Starting point & SmolLM3-3B-Base & \multicolumn{3}{c}{$\pi_{\mathrm{mixsft}}$} \\
Algorithm & Cross-entropy & \multicolumn{3}{c}{GRPO} \\
\midrule
train batch size & 128 & 128 & 128 & 128 \\
mini batch size & --- & 32 & 32 & 32 \\
Learning rate & $4\times10^{-5}$ & \multicolumn{3}{c}{$1\times10^{-6}$} \\
lr schedule & cosine, warmup 3\% & \multicolumn{3}{c}{constant, warmup 10 steps} \\
grad clip & 0.2 & \multicolumn{3}{c}{1.0} \\
\midrule
Prompt limit & --- & 1{,}024 & 2{,}048 & 2{,}048 \\
Response limit & 32{,}768 (total length) & 30{,}000 & 30{,}000 & 2{,}048 \\
rollout $n$ & --- & 16 & 16 & 8 \\
Temperature & --- & \multicolumn{3}{c}{1.0} \\
\midrule
epochs / steps & 4 epochs & 1{,}000 steps & 1{,}000 steps & 3{,}000 steps \\
$\mathrm{clip}_{\mathrm{low}}/\mathrm{clip}_{\mathrm{high}}$ & --- & \multicolumn{3}{c}{0.2 / 0.25} \\
KL penalty & --- & \multicolumn{3}{c}{disabled} \\
entropy coeff & --- & \multicolumn{3}{c}{0} \\
group filtering & --- & \multicolumn{3}{c}{by acc, at most 8 generation batches} \\
\bottomrule
\end{tabular}
\end{table}

\paragraph{Single-domain OPD and multi-teacher OPD.}
The two share the same optimiser and distillation settings, so the difference between the RouteOPD
oracle and M-OPD in the main text comes only from the number of teachers and the domain mixture, not
from the training configuration. The dense reward is computed on the student top-$k$ ($k{=}16$), and
$K{=}\text{train batch}/\text{mini batch}{=}4$ is the off-policy depth.
The multi-teacher column additionally enables the domain sampler and the three mechanism switches:
token-share balancing sets the target gradient share, gap-following allocation adjusts the domain
weights by the remaining gap, and reward refresh refreshes the dense reward within every inner
update; the table gives their values in the final recipe.

\begin{table}[t]
\caption{Hyperparameters of stage three. The left columns are the three single-domain OPD runs (the
RouteOPD oracle of the main text) and the right column is multi-teacher OPD.
The two share the optimiser, sequence lengths and distillation settings; the token-share balancing /
gap-following allocation / reward refresh rows are the values of our mechanisms in the final recipe.}
\label{tab:app_hp_opd_zh}
\centering
\small
\setlength{\tabcolsep}{3.5pt}
\resizebox{\textwidth}{!}{%
\begin{tabular}{@{}p{0.28\linewidth} ccc c@{}}
\toprule
& \multicolumn{3}{c}{Single-domain OPD (RouteOPD oracle)} & Multi-teacher OPD \\
\cmidrule(lr){2-4}\cmidrule(lr){5-5}
& math & code & IF & Open-MOPD \\
\midrule
World size & 1$\times$8 & 1$\times$8 & 1$\times$8 & 1$\times$8 \\
Teacher & $\pi_{\phi_{\mathrm{math}}}$ & $\pi_{\phi_{\mathrm{code}}}$ & $\pi_{\phi_{\mathrm{IF}}}$ & 3-teacher hard routing \\
Student starting point & \multicolumn{4}{c}{$\pi_{\mathrm{mixsft}}$} \\
\midrule
train batch size & \multicolumn{4}{c}{1{,}024} \\
mini batch size & \multicolumn{4}{c}{256 (i.e. $K{=}4$)} \\
Learning rate & \multicolumn{4}{c}{$1.5\times10^{-6}$, constant} \\
$\mathrm{clip}_{\mathrm{low}}/\mathrm{clip}_{\mathrm{high}}$ & \multicolumn{4}{c}{0.2 / 0.28} \\
KL penalty & \multicolumn{4}{c}{disabled} \\
Advantage estimation & \multicolumn{4}{c}{dense reward written directly into the advantage slot} \\
\midrule
Prompt limit & 1{,}024 & 2{,}048 & 2{,}048 & 2{,}048 \\
Response limit & 16{,}384 & 16{,}384 & 1{,}024 & 16K (math/code) / 2K (IF) \\
Total steps & \multicolumn{4}{c}{600} \\
\midrule
Reward form & \multicolumn{3}{c}{single-teacher dense reward} & multi-teacher hard-routed dense reward \\
top-$k$ & \multicolumn{4}{c}{16, from the student distribution, nucleus truncation $p{=}0.99$} \\
Reward weighting & \multicolumn{4}{c}{weighted by student probability $\tilde\pi_\theta$ (Eq.~\ref{eq:opd_reward_zh})} \\
\midrule
% Domain sampling & --- & --- & --- & weighted sampling, math:code:IF $=2{:}2{:}1$ \\
share target $g^\star$ & --- & --- & --- & $(1/3,1/3,1/3)$ \\
gap-following $\alpha$ & --- & --- & --- & $1.0$ ($m_d/m_{\mathrm{ref}}$ forward) \\
reward refresh & \multicolumn{3}{c}{enabled} & enabled \\
\bottomrule
\end{tabular}%
}
\end{table}

\end{document}